%% file: thesis.tex
\documentclass[11pt, a4paper, titlepage]{memoir}

\usepackage{graphicx}
\usepackage{amsmath}
\usepackage{amsthm}
\usepackage{amssymb}
\usepackage{makecell}
\usepackage{url}
\usepackage{tabularx,ragged2e,booktabs,caption}
\usepackage{changepage}
\usepackage{subcaption}
\usepackage{algorithm}
\usepackage[noend]{algpseudocode}
\usepackage{listings}
\usepackage[utf8]{inputenc}
\usepackage{grffile}
\usepackage{textcomp}

\usepackage{pdfpages}

\usepackage[linkcolor=black,colorlinks=true,citecolor=black,filecolor=black]{hyperref}

\makeatletter
\def\BState{\State\hskip-\ALG@thistlm}
\makeatother

\input{commands}

\input{layoutsetup}

\title{Hyperbolic sentence representations for solving Textual Entailment}
\author{Petrovski Igor}
\thesistype{Master Thesis}
\advisors{\textbf{Supervisor:}\\ Prof.\ Dr.\ Thomas Hofmann\\ \hfill \break \textbf{Co-supervisors:}\\ Octavian Ganea\\ Gary Bécigneul}
\department{\hfill \break Department of Computer Science}
\date{April 24, 2018}

\begin{document}

\frontmatter

\begin{titlingpage}
  \calccentering{\unitlength}
  \begin{adjustwidth*}{\unitlength-24pt}{-\unitlength-24pt}
    \maketitle
  \end{adjustwidth*}
\end{titlingpage}

\include{abstract}
\newpage
\renewcommand{\abstractname}{Acknowledgements}
\include{acknowledgements}
\newpage
\tableofcontents

\mainmatter
\input{introduction.tex}
\input{baselines.tex}

\input{core_chapter.tex}

\input{datasets.tex}

\input{experiments_chapter.tex}

\input{conclusion.tex}

\bibliography{references}
\bibliographystyle{ieeetr}

\end{document}

%% file: commands.tex
\newcommand{\R}{\mathbb{R}}
\newcommand{\B}{\mathcal{B}}
\newcommand{\M}{\mathcal{M}}
\newcommand{\D}{\mathcal{D}}
\newcommand{\Loss}{\mathcal{L}}
\newcommand{\norm}[1]{\left\lVert#1\right\rVert}
\DeclareMathOperator*{\argmin}{argmin} 

%% file: layoutsetup.tex
\usepackage{ETHlogo}

\nonzeroparskip
\defaultlists

\makeatletter

\makepagestyle{ruledCentered}
\if@twoside
\copypagestyle{ruledCentered}{Ruled}
\else
\copypagestyle{ruledCentered}{ruled}
\fi
\makeoddfoot{ruledCentered}{}{\thepage}{}
\makeevenfoot{ruledCentered}{}{\thepage}{}
\copypagestyle{chapter}{ruledCentered}

\makeoddhead{chapter}{}{}{}
\makeevenhead{chapter}{}{}{}
\makeheadrule{chapter}{\textwidth}{0pt}
\copypagestyle{acknowledgements}{empty}
\copypagestyle{abstract}{empty}

\setlrmargins{*}{*}{1}

\makechapterstyle{bianchimod}{%
  \chapterstyle{default}
  \renewcommand*{\chapnamefont}{\normalfont\Large\sffamily}
  
  \renewcommand*{\printchaptername}{%
    \chapnamefont\centering\@chapapp}

  }

\chapterstyle{bianchimod}

\setsecheadstyle{\Large\bfseries\sffamily}
\setsubsecheadstyle{\large\bfseries\sffamily}
\setsubsubsecheadstyle{\bfseries\sffamily}
\setparaheadstyle{\normalsize\bfseries\sffamily}
\setsubparaheadstyle{\normalsize\itshape\sffamily}
\setsubparaindent{0pt}

\captionnamefont{\sffamily\bfseries\footnotesize}
\captiontitlefont{\sffamily\footnotesize}
\setsecnumdepth{subsection}
\settocdepth{subsection}

\pretitle{\vspace{0pt plus 0.7fill}\begin{center}\HUGE\sffamily\bfseries}
\posttitle{\end{center}\par}
\preauthor{\par\begin{center}\let\and\\\Large\sffamily}
\postauthor{\end{center}}
\predate{\par\begin{center}\Large\sffamily}
\postdate{\end{center}}

\def\@advisors{}
\newcommand{\advisors}[1]{\def\@advisors{#1}}
\def\@department{}
\newcommand{\department}[1]{\def\@department{#1}}
\def\@thesistype{}
\newcommand{\thesistype}[1]{\def\@thesistype{#1}}

\renewcommand{\maketitlehookb}{\vspace{1in}%
  \par\begin{center}\Large\sffamily\@thesistype\end{center}}

\renewcommand{\maketitlehookd}{%
  \vfill\par
  \begin{flushright}
    \sffamily
    \@advisors\par
    \@department, ETH Z\"urich
  \end{flushright}
}

\checkandfixthelayout

\makeatother


%% file: abstract.tex
\begin{abstract}
Hyperbolic spaces have proven to be suitable for modeling data of hierarchical nature. As such we use the Poincare ball to embed sentences with the goal of proving how hyperbolic spaces can be used for solving Textual Entailment. To this end, apart from the standard datasets used for evaluating textual entailment, we developed two additional datasets. We evaluate against baselines of various backgrounds, including LSTMs, Order Embeddings and Euclidean Averaging, which comes as a natural counterpart to representing sentences into the Euclidean space. We consistently outperform the baselines on the SICK dataset and are second only to Order Embeddings on the SNLI dataset, for the binary classification version of the entailment task.
\end{abstract}

%% file: acknowledgements.tex
\begin{abstract}

I would like to thank professor Thomas Hofmann for the opportunity to do my master thesis under his supervision. I am extremely thankful to my co-supervisors Octavian Ganea and Gary Bécigneul for their responsiveness, advices and constant support.

I will be forever grateful to my parents, my sister and my girlfriend for their enormous support throughout my master studies and being always there for me.

\end{abstract}

%% file: introduction.tex

\chapter{Introduction}
\label{ch:introduction}

Recognizing Textual Entailment or \textbf{RTE} is one of the fundamental problems in the field of Natural Language Understanding. It represents a classification problem of pair of sentences on the basis of whether one of the sentences can be inferred by the other. In Table \ref{table:entailment_examples} we can see some entailment examples.

\begin{center} 
	\begin{adjustwidth}{-1.5cm}{}
	\begin{tabular}{|c|c|}
	\hline
	\textbf{Premise} & \textbf{Hypothesis}\\
	\hline
	\makecell{A senior is waiting at \\ the window of a restaurant that serves sandwiches.}
	&
	A person waits to be served his food . \\
	\hline
	Two doctors perform surgery on patient .        
	&
	Doctors are performing surgery . \\
	\hline
	\makecell{A group of people standing in the snow  \\ with a mountain in the background .}     
	&
	People are outside . \\
	\hline
	\end{tabular}
	\end{adjustwidth}
	\label{table:entailment_examples}
	\captionof{table}{Pairs of sentences that represent entailment taken from the SNLI corpus \cite{snli_corpus}}
\end{center}

Solving this problem has vast applicabilty to a variaty of tasks. The ability to be able to distinguish the relation between pairs of sentences and whether one of the sentences is a more general (or specific) version of the other can be very useful in enhancing Question-Answering platforms. Furthermore, distinguishing the degree to which one sentence is a generalization of the other can be extremely helpful to constructing knowledge datasets and redundancy detection.

\section{Motivation}

Distributed continuous word embeddings have become the most efficient way of representing words and have seen huge applicability in text handling. Using the Skip-Gram method introduced in \cite{skipgram} for obtaining word embeddings in an unsupervised manner has proven to be very effective. A natural extension to the Skip-Gram method is the Skip-Thoughts method for representing sentences in a continuous vector space as shown in \cite{skipthoughts}. In a similar unsupervised manner to embedding words, the sentences are embedded as points in the Euclidean vector space. As we can see from \cite{a_simple_but_tough} and \cite{fast_sent_jaggi} there are different approaches to embedding sentences in the Euclidean space, with varying degrees of success.

In this work, we present a completely novel approach to embedding sentences as points in a hyperbolic space. By exploiting the features of the hyperbolic space we try a different approach to solving the problem of Textual Entailment. To the best of our knowledge, we are the first to present a sentence representation method in a hyperbolic space.

\section{Thesis contribution}

The contribution of the work done can be summed up in the following points:
\begin{itemize}
    \item We provide a completely novel approach to sentence embeddings by utilizing the hyperbolic space along with an explanation of the theoretical background of the said approach. 
    \item We contribute by creating 2 datasets used to analyze some of the features of the mentioned model.
    \item Python implementation using the Tensorflow framework \cite{tensorflow}.
\end{itemize}

\section{Outline}
In Chapter \ref{ch:related_work} we talk about related work and approaches taken to tackle the problem of textual entailment. We put an emphasis on the baselines we compare against and explain them in detail. After that, in Chapter \ref{ch:core} we talk about hyperbolic spaces, poincare embeddings, riemannian optimization and we introduce our model along with describing its complexities. Next, in Chapter \ref{ch:datasets}, we outline the datasets that we use throughout the experiments and point out some of their key features and characteristics. In Chapter \ref{ch:experiments} we present the experiments done and provide analysis of our model's behavior. Finally, in Chapter \ref{ch:conclusion}, we provide a conclusion to the work done and how it can be applied to other tasks pertaining to sentence representations.


%% file: baselines.tex


\chapter{Related Work}
\label{ch:related_work}

In this section we present our  baselines that we compare against our newly introduced models. Each one of them is picked in such a way that it addresses certain aspects of our model's features.  We compare against three baselines: Euclidean averaging, LSTMs and Order embeddings. Every one of the models approaches the problem of textual entailment from a different angle and is related to our model in a specific way. Euclidean averaging is the natural counterpart to mobius summation and mobius averaging in the euclidean space, where we have the exact same number of trainable model parameters. LSTMs represents the classical approach to tackling textual entailment with encompassing the context through its units. Order embeddings, on the other hand, have proven to be very effective in exploiting the hierarchical nature of the textual-entailment problem and its approach of introducing ordered pairs proves to be quite effective and successful. We briefly describe each of the aforementioned baselines.

\section{Euclidean averaging} \label{section:euclidean_averaging}

Euclidean averaging, as a sentence representation method, is a method where we simply sum up the word embeddings and divide by the number of tokens in the sentence. Although the method is of a very simplistic nature, it proved to be a good baseline to compare against. Even though it loses the order of the words in the sentence it is still a powerful method for textual entailment, when the word embeddings are specifically trained for this task. This approach was explored in further depth in \cite{siamese}. The sentence representation is defined as: 

\begin{equation}
    s = \frac{1}{N}\sum_{i = 1}^{N}w_i
\end{equation}

where $N$ is the number of words in the sentence and $w_{1..N}$ are its constituent words. This method, apart from the parameters in the FFNN and the word embeddings doesn't require any additional trainable parameters, which makes it very fast to train. As with the LSTMs model, that is introduced in the next section, we use the standard model architecture for textual entailment that is shown in Figure \ref{fig:general_architecture}.

\section{LSTMs} \label{section:lstm}
Recurrent Neural Networks are widely used in text understanding problems. Given their recurrent nature, they are a natural method to tackling problems like language modeling, question-answering,  sentiment classification and textual entailment. RNNs for language modeling were introduced in \cite{rnn_language_models}.

Unlike methods like n-grams or euclidean averaging, recurrent neural networks try to embed the context of the sequence they are processing into a $d$-dimensional hidden state vector. Using this hidden state vector the model is able to maintain some sort of a context when processing the next element in the sequence. In the following equation we can see how the hidden state at time $t$ is calculated.

\begin{equation} \label{eq:rnn}
	h_t = f(h_{t-1}, x_t; \theta)
\end{equation}

As we can see from Equation \ref{eq:rnn} each hidden state at step $t$ depends on the previous hidden state $h_{t-1}$ and the current input $x_t$ that is being processed. In Figure \ref{fig:rnn} we can see a visualization of an RNN.

\begin{figure}
\begin{subfigure}{.45\textwidth}
  \centering
  \includegraphics[width=.85\linewidth]{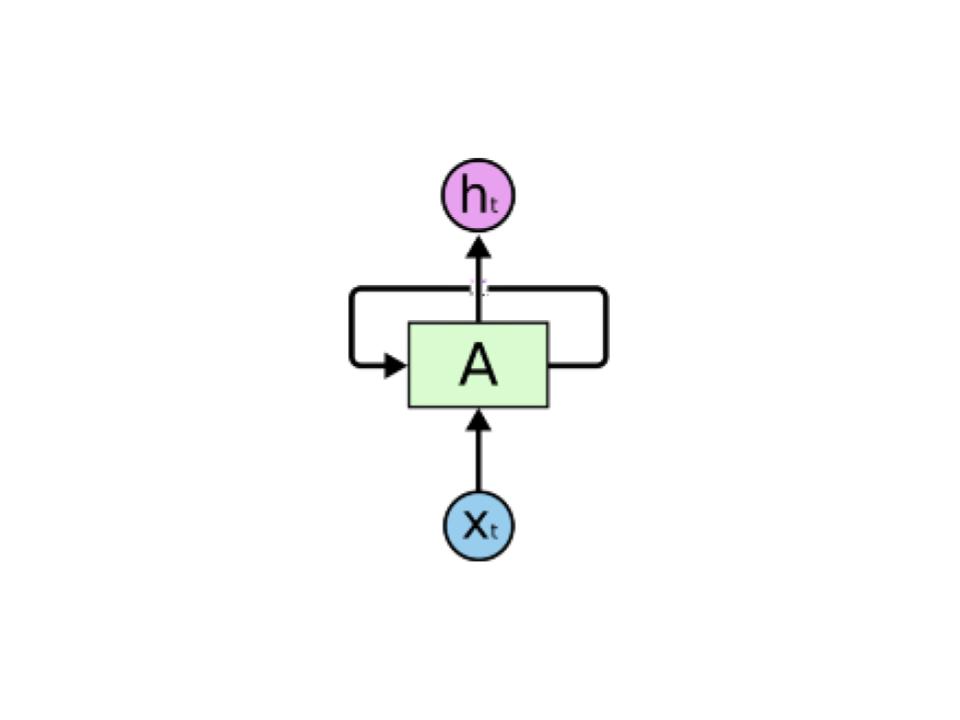}
  \caption{RNN}
  \label{fig:rnn}
\end{subfigure}%
\hspace{0.1\textwidth}
\begin{subfigure}{.45\textwidth}
  \centering
  \includegraphics[width=.85\linewidth]{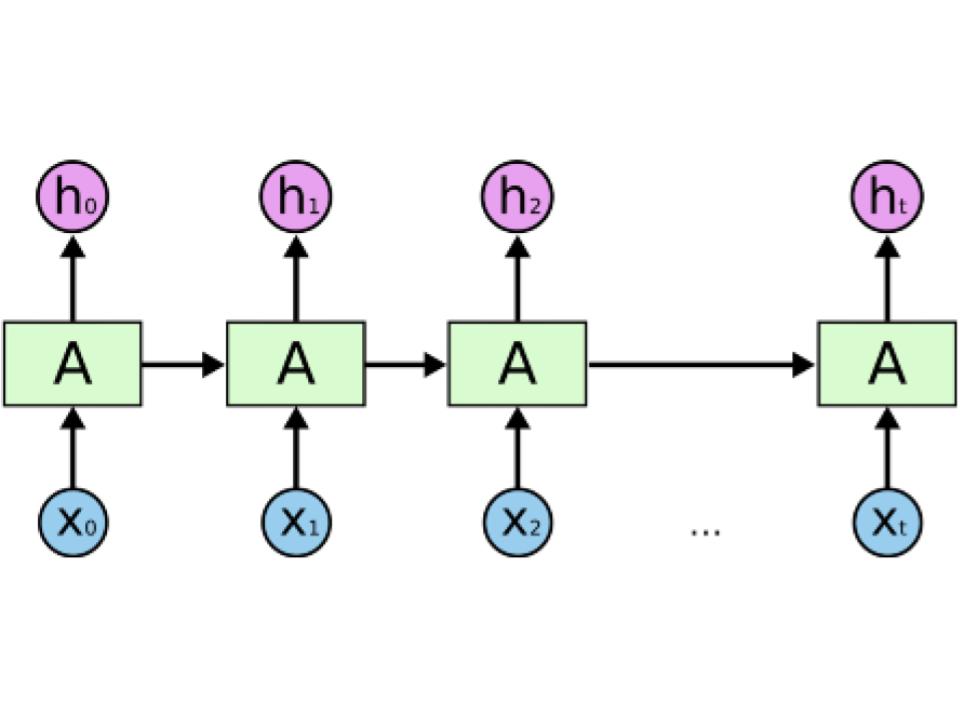}
  \caption{Unrolled version of an RNN}
  \label{fig:rnn_unrolled}
\end{subfigure}
\caption{RNN visualization \cite{rnn_blog}}
\label{fig:rnn}
\end{figure}

Although theoretically sound, this vanilla RNN model suffers from the vanishing gradient problem. When the sequences get relatively long, the backpropagation error that gets propagated through time diminishes and isn't able to update the parameters of the model properly. One of the solution to this problem is introducing Long Short Term Memory \cite{lstms} cells that deal with this problem. This is done by introducing gate units defined by: 

\begin{equation}
	f_t = \sigma \left( W_{f,1} h_{t-1} + W_{f,2} x_t + b_f \right)
\end{equation}

\begin{equation}
	i_t = \sigma \left( W_{i,1} h_{t-1} + W_{i,2} x_t + b_i \right)
\end{equation}

\begin{equation}
	\widetilde{c}_t 	= tanh \left( W_{c,1}h_{t-1} + W_{c,2}x_t + b_c \right)
\end{equation}

\begin{equation}
	c_t = f_t \odot c_{t-1} + i_t \odot \widetilde{c}_t
\end{equation}

\begin{equation}
	o_t = \sigma \left( W_{o,1} h_{t-1} + W_{o, 2} x_t + b_o \right)
\end{equation}

\begin{equation}
	h_t = o_t tanh \odot \left( c_t \right)
\end{equation}
where $W_{f,1}, W_{i,1}, W_{c,1}, W_{o,1} \in \R^{h \times h}$ , $W_{f,2}, W_{i,2}, W_{c,2}, W_{o,2} \in \R^{h \times m}$ , $f_{t}, i_{t}, \widetilde{c}_{t}, c_{t}, o_t, h_t \in \R^{h}$ and $\odot$ represents point-wise multiplication.

LSTMs have been proven to be one of the standard approaches to encoding sentences for various tasks. Even though there are multiple versions of LSTMs that address the problem of textual entailment, as shown in \cite{rocktaschel2015reasoning}, \cite{wang2015learning} and \cite{liu2016learning}, here we focus on the basic version. 

\begin{figure}[h!]
 \begin{center}
 \includegraphics[scale=0.3]{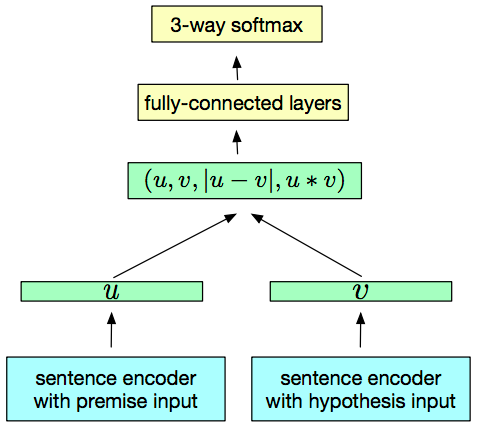}
 \end{center}
 \caption{General architecture for Textual Entailment \cite{nli_learning_scheme}}
 \label{fig:general_architecture}
\end{figure}

Having defined LSTMs, we use them as sentence encoders to obtain representations for both the premise and the hypothesis when solving the textual entailment problem.  As it can be seen from Figure \ref{fig:general_architecture}, after obtaining the sentence representations we use various concatenation methods as an input to the Feed Forward Neural Network that uses softmax for classification. All the parameters in the model are updated through backpropagation.

\section{Order embeddings}\label{section:order_embeddings}
We also compare against the baseline of Order embeddings introduced in \cite{order_embeddings}. The idea is to introduce order between objects by introducing partial order completion where we are given a set of positive samples of ordered pairs $P = {(u,v)}$ and a set of negative samples of unordered pairs $N$. Based on this training dataset our goal is to predict whether an unseen pair $(x, y)$ is ordered or not. The loss function used to train the embeddings is as follows:

\begin{equation}
L = \sum_{(p,h) \in P}E(f(p), f(h)) + \sum_{(p',h') \in N} max\{0, \alpha - E(f(p'), f(h'))\}
\end{equation}

where the loss function is a composition of positive samples - pairs of sentences that represent entailment and negative samples - sentences that don't. $f(p)$ represents the function that is used to obtain the sentence representation from sentence $p$, whereas $E(x,y)$ is the score function defined as:

\begin{equation}
    E(x, y) = \norm{max\{0, y - x\}}^{2}
\end{equation}

which encourages objects that are higher up the hierarchy to be closer to the origin. The score function is defined in such manner so that it satisfies the following property of order entailment:

\begin{equation}
x \preceq y \hspace{0.3cm} \text{if and only if} \hspace{0.3cm} \bigwedge_{i = 1}^{D} x_i \geq y_i
\end{equation}

where $D$ is the number of dimensions.


%% file: core_chapter.tex

\chapter{Models}
\label{ch:core}

\section{Introduction}
In this chapter we introduce the models we have developed for representing sentences in a hyperbolic space. We first talk about hyperoblic spaces in general and why we find them useful and suitable for our task. We introduce some basic concepts that are useful to know in order to grasp the overall model architecture. Then, we address the word embedding method in hyperbolic space that has been introduced in \cite{poincare}. After that, we talk about Mobius addition as a composition method between two points in the poincare ball. Using mobius addition, we next present the algorithm for obtaining sentence representations, given their parse trees. Next, we introduce the models that have been developed in this work and we talk about Riemannian optimization and Riemannian Stochastic Gradient Descent as a way to train the model parameters. 

Before delving further into depth about Hyperbolic spaces, we introduce some basic concepts about differential geometry. 

\subsection{Manifold}
An n-dimensional manifold $\M_n$ is a topological space such that each point in that manifold is locally homeomorphic\footnote{Two spaces are homeomorphic if there is a mapping between those spaces such that all of the topological properties are preserved} to an n-dimensional Euclidean space. Even though every point in a manifold has a local topological neighborhood  that is homeomorphic to a Euclidean space, globally manifolds are not homeomorphic to the Euclidean space.
\subsection{Tangent space}
\begin{figure}[h!]
 \begin{center}
 \includegraphics[scale=0.3]{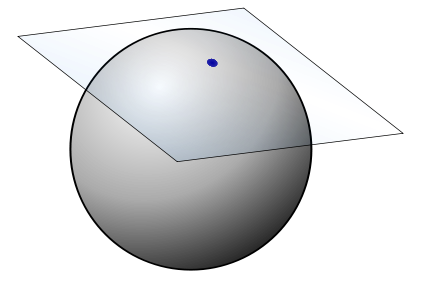}
 \end{center}
 \caption{Tangent space of a point \cite{image_tangent_space}}
 \label{fig:tangent_space}
\end{figure}

Let $x$ be a point in a manifold $\M_n$.  Intuitively speaking, if we attach an $\R^{n}$  space to $x$ such that the space is tangential to $\M_n$, we get the tangent space of $x$ denoted as  $T_x\M$. In Figure \ref{fig:tangent_space} we can see the tangent space of a point. The elements of  $T_x\M$ are called tangent vectors at x.

\subsection{Riemannian manifold}
In order to define what Riemannian manifold is, we first need to introduce the Riemannian metric tensor. Let $u_x$ and $v_x$ be tangent vectors of the tangent space $T_x\M$ at point $x$ $\in$ $\M_n$. A function $g(u_x, v_x)$ that takes the tangent vectors $u_x, v_x$ as input and computes a real number such that it generalizes most of the properties of a dot product in a Euclidean space, is called a metric tensor. Riemannian manifold is a smooth manifold which is equipped with a positive definite\footnote{$g(u,u) > 0$ for every non-zero vector $u$} metric tensor $g$. The Riemannian metric $g_x: T_x\M \times T_x\M \rightarrow \R$ represents a group of inner-products defined over the tangent spaces of every point $x$ in a manifold $\M$. It gives the infinitesimal distance in a manifold and thus it is used in defining the distance between any two points in a manifold.

\subsection{Geodesics}
A geodesic between two points is a locally shortest smooth curve that connects those points. Geodesics can be seen as a generalization of the concept of straight lines to manifolds. 
\section{Hyperbolic spaces}
\begin{figure}[h!]
 \begin{center}
 \includegraphics[scale=0.3]{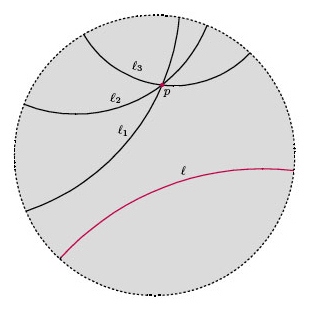}
 \end{center}
 \caption{Parallel postulate doesn't hold in the hyperbolic geometry \cite{img_parallel_postulate}}
 \label{fig:parallel_postulate}
\end{figure}

The hyperbolic space is a simply connected\footnote{A space is simply connected if there is a path between any two points and any path can be continuously shrinked into a point while remaining in the domain} Riemannian manifold with constant negative curvature. Another way of looking at it is that the hyperbolic space is a geometrical space analogous to the euclidean space, with the difference being that the parallel postulate doesn't hold in the hyperbolic space. As we can see from Figure \ref{fig:parallel_postulate} there are infinitely many lines parallel to line $l$ passing through point $p$ i.e. the parallel postulate doesn't hold.\\

One of the distinctive properties of the hyperbolic space is that when it is embedded in a Euclidean space, every point in the hyperbolic space is a saddle point. Another property, very relevant to the work done in this thesis, is that the amount of space covered by the n-ball in the n-dimensional hyperbolic space increases exponentailly with the size of the ball's radius. This is not the case with a ball embedded in a Euclidean space, which increases polynomially with respect to its radius, rather than exponentially.

This useful property makes hyperbolic spaces attractive to modeling data of hierarchical nature. Since hierarchical structures expand exponentially with respect to the depth, it is only logical to model them in a hyperbolic space which also expands exponentially with the size of its radius. This intuition has proven to bring huge success in embedding trees and networks in the hyperbolic space as shown in \cite{kleinberg2007geographic}, \cite{krioukov2010hyperbolic} and \cite{boguna2010sustaining}.

Our intuition lead us to believe that hyperbolic spaces can also be very suitable for modeling sentences, given their hierarchical structure. For this we needed sentences' parse trees so that we are able to express the sentence as a hierarchical composition of its respective constituent parts.

There are multiple models in the hyperbolic space: Klein model, hyperboloid model, Poincare ball model and Poincare half space model. It is important to note that any of the models can be transformed to any other of the models by a transformation that preserves all the properties of the space. In the work done in this thesis  we focus on the Poincare ball model which has a differentiable continuous distance function.

\section{Poincare embeddings}
The poincare unit ball is a model of hyperbolic geometry in which all of the points reside inside the ball and the boundary of the ball represents infinitely distant points. Formally, the Poincare ball model is defined as

\begin{equation}
	\B^{d} = \{ x \in \R^{d} | \norm{x} < 1 \}
\end{equation}

where $\norm{x}$ represents the norm of $x$ and $d$ is the dimensionality of the unit ball.

The Poincare ball model corresponds to the Riemannian manifold $(\B^{d}, g_x)$
equipped with the Riemannian metric tensor:

\begin{equation}
    g_x = \left( \frac{2}{1 - \norm{x}^{2}} \right)^{2}g^{E}
\end{equation}

where $x \in B^{d}$ and $g^{E}$ is the euclidean metric tensor. The distance between two points $u,v \in B^{d}$ is represented as:

\begin{equation} \label{eq:poincare_distance}
    d(u,v) = arcosh\left(1 + 2\frac{\norm{u-v}^{2}}{(1 - \norm{u}^{2})(1 - \norm{v}^{2})} \right)
\end{equation}

\begin{figure}
\begin{subfigure}{.331\textwidth}
  \centering
  \includegraphics[width=.8\linewidth]{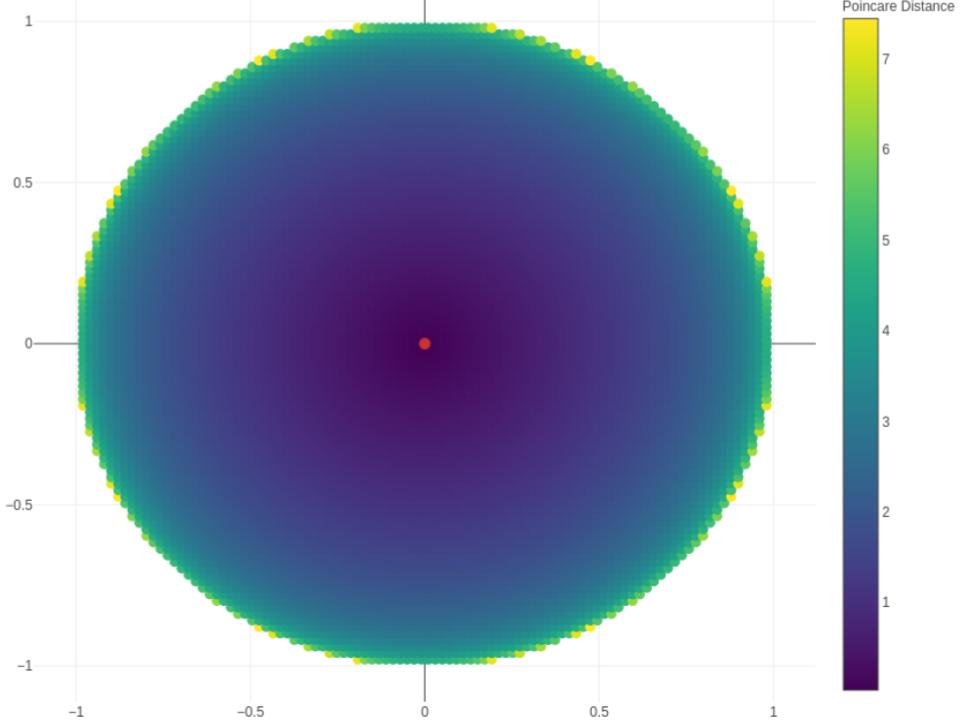}
  \caption{Point (0.0, 0.0)}
  \label{fig:sfig_poincare_dist_1}
\end{subfigure}%
\begin{subfigure}{.331\textwidth}
  \centering
  \includegraphics[width=.8\linewidth]{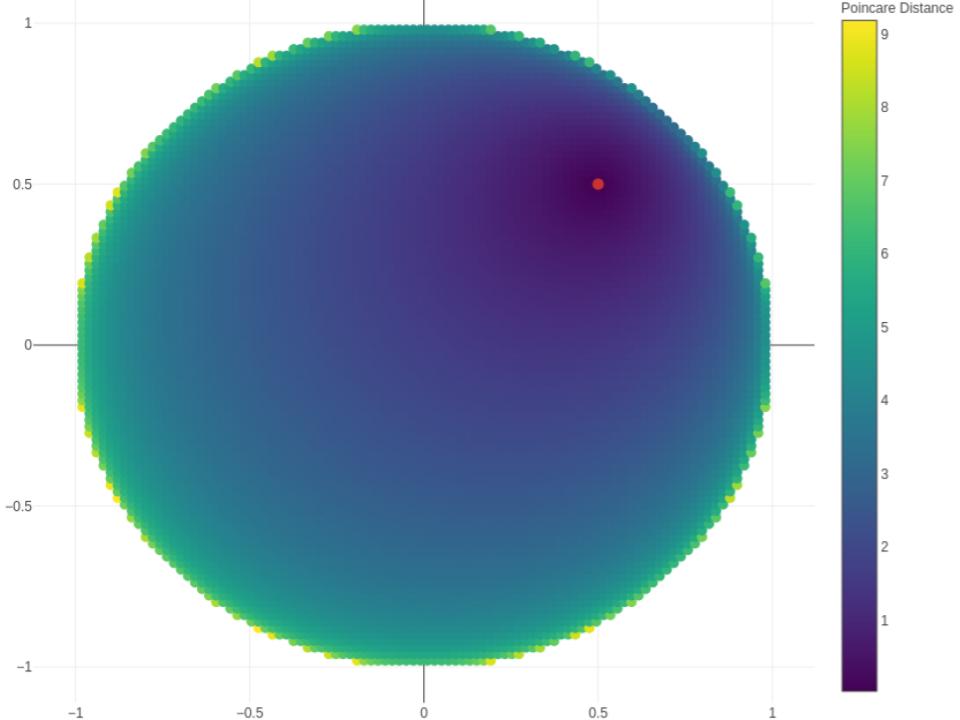}
  \caption{Point (0.5, 0.5)}
  \label{fig:sfig_poincare_dist_2}
\end{subfigure}
\begin{subfigure}{.331\textwidth}
  \centering
  \includegraphics[width=.8\linewidth]{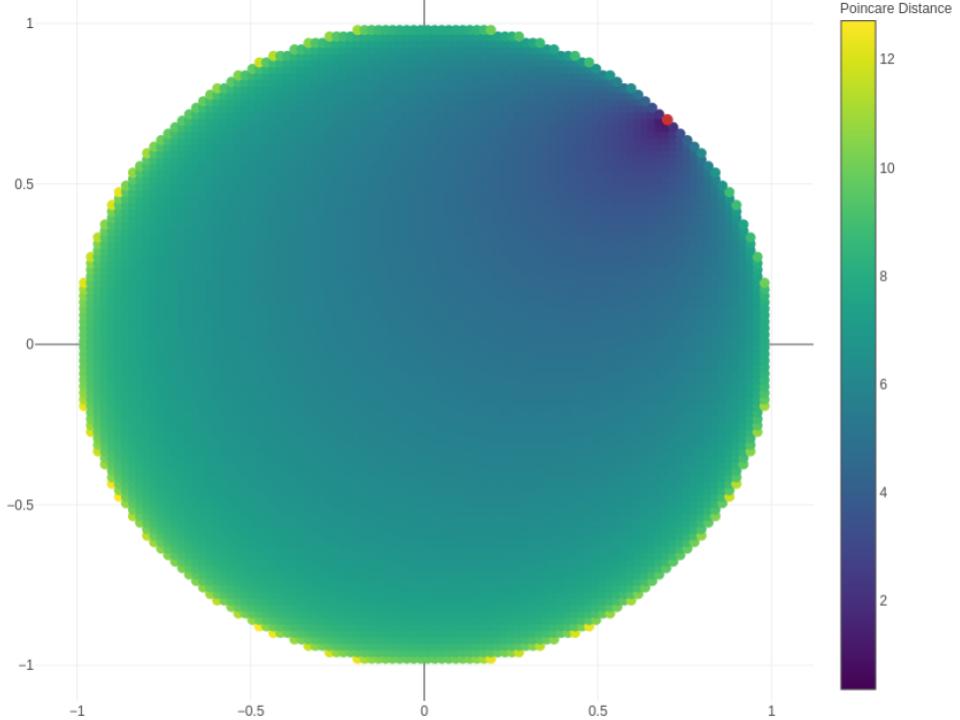}
  \caption{Point (0.7,0.7)}
  \label{fig:sfig_poincare_dist_3}
\end{subfigure}%
\caption{Poincare distance proximity \cite{poincare_dist_images}}
\label{fig:poincare_distances}
\end{figure}

We can see that the poincare distance is symmetric and that it changes smoothly with respect to the points $u$ and $v$. The smoothness makes it suitable for representing continuous word embeddings.  
In Figure \ref{fig:poincare_distances} we can see a visualization of the poincare distances for 3 different points. The darker area represents closer points, whereas the lighter area represents more distant points. Intuitivelly, we can conclude that two points are further apart, the further they are from the origin. Another good feature of the poincare distance is that it is differentiable which makes it suitable for gradient optimization methods like Riemannian Stochastic Gradient descent which we talk about in Section \ref{section:Optimization}

\section{Mobius addition as a method for node composition}
Since our goal is to come up with a sentence representation, given its constituent words and parse tree, we need a composition method that will combine the leaves and the intermediate nodes all the way up to the root of the parse tree. We use mobius addition that is defined in \cite{mobius} as a composition method of two points. We use the notation $\oplus_M$ to represent a mobius addition between two points in a Poincare unit ball and it is defined as:  
\begin{equation} \label{eq:mobius_addition}
  u \oplus_M v = \frac{(1 + 2 \langle u , v \rangle + \norm{v}^{2})u + (1 - \norm{u}^{2})v}{1 + 2 \langle u , v \rangle  + \norm{u}^{2}\norm{v}^{2}}
\end{equation}

where $\langle \cdot \rangle$ and $\norm{\cdot}$ represent the inner product and the norm of a vector, respectively.

Unlike addition in the Euclidean space, mobius addition is noncommutative and nonassociative. Intuitively, this is good because word order in sentences is important and we wouldn't like to obtain the same result regardless of the order and the structure of the sentence tree. The mobius addition is subject to some identities:

\begin{align}
	d(u,v) &= 2\tanh^{-1}\norm{-u \oplus_M v } \\
	-a \oplus_M \left(a \oplus_M b \right) &= b \\
	-a \oplus_M a &= 0 \\
	a \oplus_M a &= \frac{2a}{1 + \norm{a}^2} 
\end{align}
where $d(u,v)$ is the poincare distance between $u$ and $v$.

\begin{figure}
\begin{subfigure}{.45\textwidth}
  \centering
  \includegraphics[width=.85\linewidth]{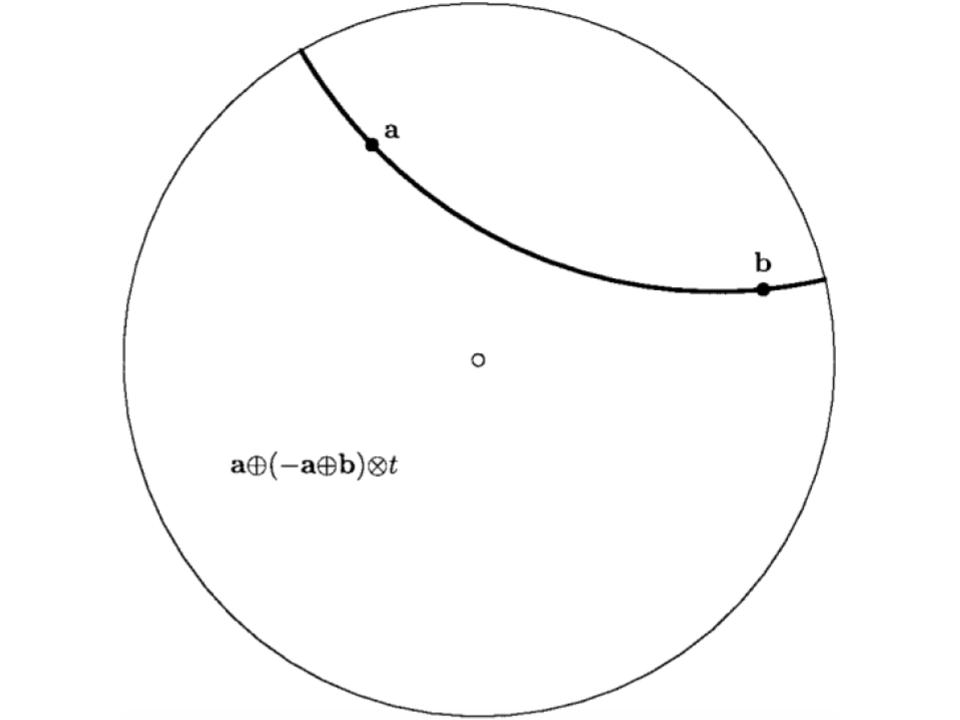}
  \caption{The geodesic passing through $a$ and  $b$ in a two-dimensional Poincare disk}
  \label{fig:sfig1}
\end{subfigure}%
\hspace{0.1\textwidth}
\begin{subfigure}{.45\textwidth}
  \centering
  \includegraphics[width=.85\linewidth]{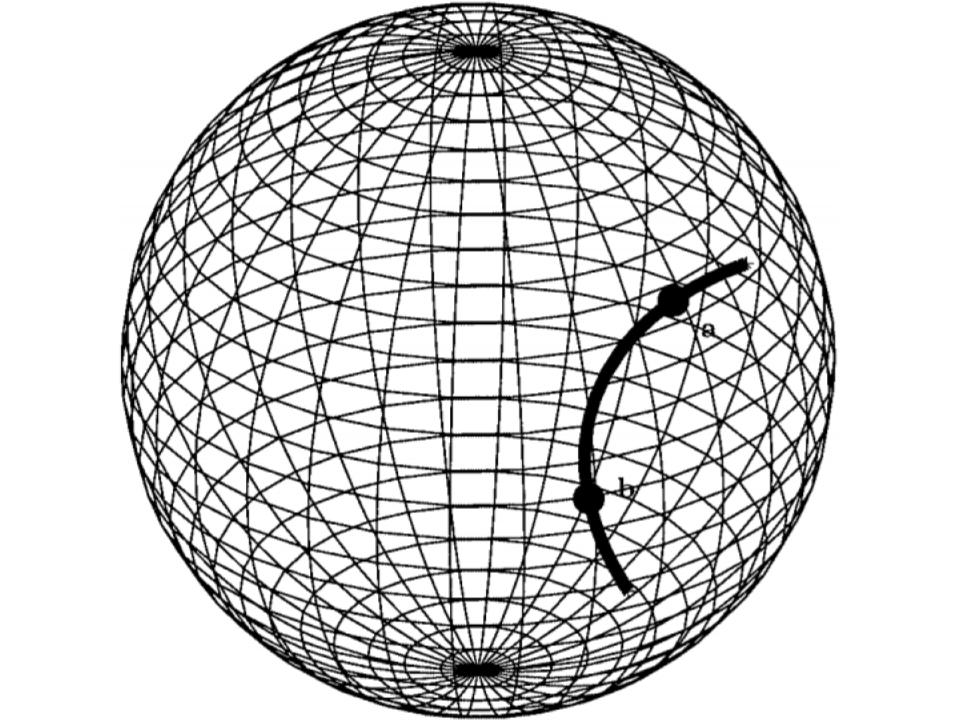}
  \caption{The geodesic passing through $a$ and $b$ in a three-dimensional Poincare ball}
  \label{fig:sfig2}
\end{subfigure}
\caption{Mobius geodesics \cite{mobius}}
\label{fig:mobius_geodesic}
\end{figure}

We use these identities throughout our code as sanity checks. Using mobius addition, the geodisics passing through any two points $a$ and $b$ can be defined as: 

\begin{equation}
	a \oplus_M \left( \ominus a \oplus_M b\right) \otimes_M t
\end{equation}

where $t \in \R$ and $\ominus a = -a$. $\otimes_M$ represents Mobius multiplication and we talk about it in Section \ref{section:mobius_averaging_ffnn}. When $t = 0$ the geodesic passes through $a$ and when $t=1$ it passes through $b$. We can see a visualization of this formula in Figure \ref{fig:mobius_geodesic}

Having defined the composition method, we now turn to explaining how we use it to come up with a sentence representation.

\section{Sentence representation} \label{section:sentence_representation}
Having defined the composition method between two points in the Poincare unit ball, we need a way to come up with a sentence embedding, given its parse tree and its constituent word embeddings. In the following pseudocode we outline the algorithm that produces the sentence representation. As input we receive the sentence's parse tree and the individual word embeddings. We then do a post order traversal of the given parse tree in order to obtain the needed structures that are used for the sentence construction. After obtaining those structures in line 9, we do another post order traversal of the tree, which is defined in the procedure BUILD. This second traversal recursively defines every internal node as a mobius addition of its respective child nodes. Following this recursive approach we obtain the root value, which we use as a sentence representation.

\begin{algorithm}
\caption{Algorithm for obtaining the sentence representation}\label{algo:sent_rep}
\begin{algorithmic}[1]
\Procedure{SentenceRepresentation($tree$, $embeddings$)}{}

\State $node\_to\_ind \gets \{\}$
\State $ind \gets 0$

\State

\For{each $node$ in $tree.nodes$}
    \State $node\_to\_ind(node) \gets ind$
    \State $ind \gets ind + 1$
\EndFor

\State

\State $is\_leaf, left, right, word\_ids \gets \textsc{PostOrderTraversal} (tree, node\_to\_ind)$
\State $root\_ind \gets \textsc{Len}(word\_ids) - 1$
\State $result \gets \textsc{Build}(is\_leaf,left,right,word\_ids,embeddings,root\_ind)$
\State
\State \Return $result$
\EndProcedure

\State

\Procedure{Build($is\_leaf$, $left$, $right$, $word\_ids$, $embeddings$, $i$)}{}
\If{$is\_leaf(i) = TRUE$}
	\State \Return $embeddings(word\_ids(i))$
\Else
    \State $l\_representation \gets \textsc{Build}(is\_leaf,left,right,word\_ids,embeddings,left(i))$ 
    \State $r\_representation \gets \textsc{Build}(is\_leaf,left,right,word\_ids,embeddings,right(i))$
    \State \Return $\textsc{MobiusAddition}(l\_representation, r\_representation)$
\EndIf
\EndProcedure

\State

\Procedure{PostOrderTraversal($tree, node\_to\_ind$)}{}
	\State $is\_left \gets \{\}$
	\State $left \gets \{\}$
	\State $right \gets \{\}$
	\State $word\_ids \gets \{\}$
	\State $\textsc{Populate}(node\_to\_ind, is\_leaf, left, right, word\_ids, tree.root)$	
	
	\State \Return $isleaf, left, right, wordids$
\EndProcedure

\State 

\Procedure{Populate($node\_to\_ind,is\_leaf, left, right, word\_ids, node$)}{}
\If{$node.left == NULL$ \textbf{and} $node.right == NULL$}
	\State $isleaf \gets is\_leaf + \{ TRUE\}$
	\State $left \gets left + \{ -1 \}$
	\State $right \gets right + \{ -1 \}$
	\State $word\_ids \gets word\_ids + \{ node.word\_id \}$
\Else
	\State $\textsc{Populate}(is\_leaf, left, right, word\_ids, node.left)$
	\State $\textsc{Populate}(is\_leaf, left, right, word\_ids, node.right)$
	\State $isleaf \gets isleaf + \{ FALSE\}$
	\State $left \gets left + \{node\_to\_ind(node.left) \}$
	\State $right \gets right + \{node\_to\_ind(node.right) \}$
	\State $word\_ids \gets word\_ids + \{ node.word\_id \}$
\EndIf
\EndProcedure

\end{algorithmic}
\end{algorithm}

Our code is loosely based on this pseudocode, although for the sake of efficiency we precalculate all the necessary representations so that we don't redo the same work during training.

\begin{figure}[h!]
 \begin{center}
 \includegraphics[scale=0.3]{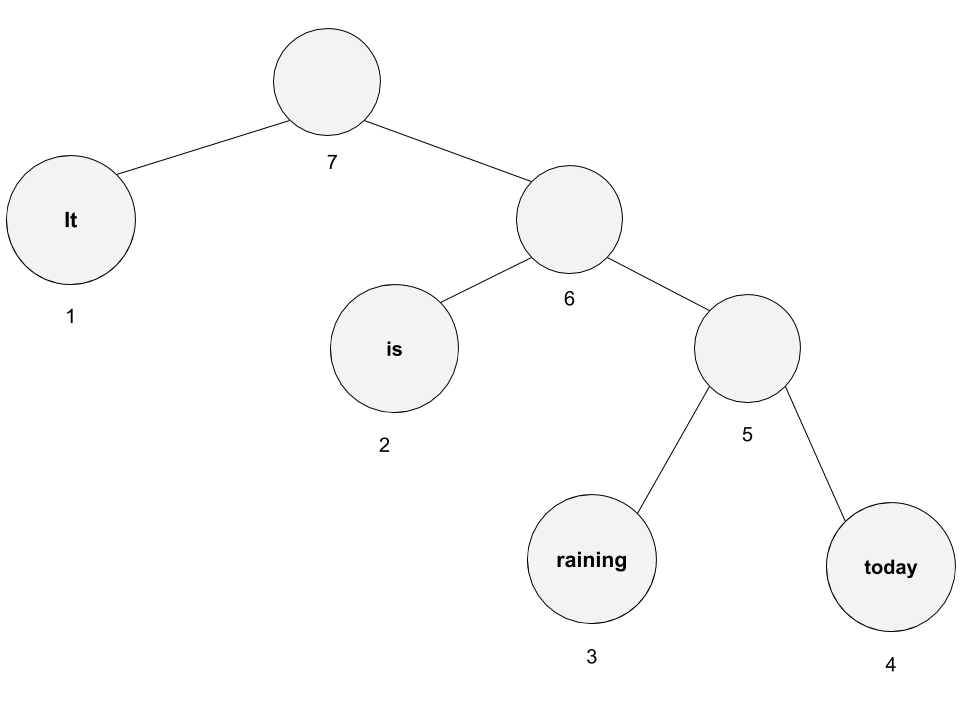}
 \end{center}
 \caption{Parse tree for the sentence "It is raining today". The numbers below the nodes represent the order of the post order traversal.}
 \label{fig:parse_tree_example}
\end{figure}

In Figure \ref{fig:parse_tree_example} we can see a parse tree example for the sentence "It is raining today" which, using the post order traversal, it is represented through mobius summation as: 

\begin{equation}
   \theta_{ROOT} = \theta_{It} \;  \oplus_{M} \; ( \theta_{is} \; \oplus_M \;  (\theta_{raining} \; \oplus_M \; \theta_{today} ) \; )  
\end{equation}

where $\theta_{w}$ represents the corresponding word embedding for the word $w$.

\begin{figure}
\begin{subfigure}{.45\textwidth}
  \centering
  \includegraphics[width=1.0\linewidth]{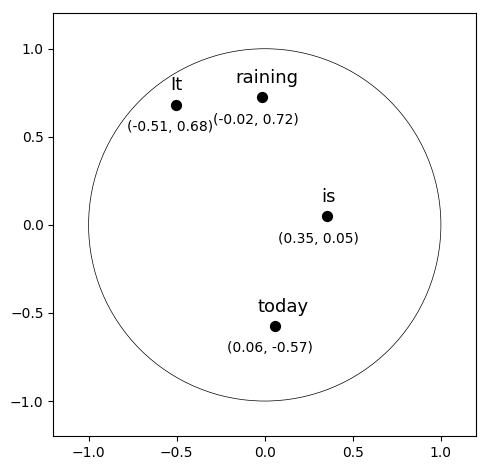}
  \caption{Initial placement of the word embeddings}
  \label{fig:mob_add_0}
\end{subfigure}%
\hspace{0.1\textwidth}
\begin{subfigure}{.45\textwidth}
  \centering
  \includegraphics[width=1.0\linewidth]{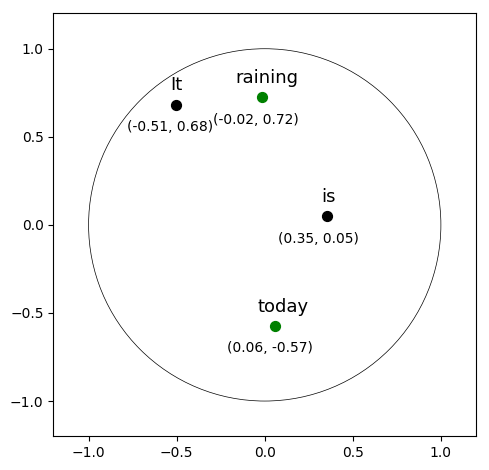}
  \caption{$\theta_{\text{raining}} \oplus_M \theta_{\text{today}}$}
  \label{ffig:mob_add_1}
\end{subfigure}

\begin{subfigure}{.45\textwidth}
  \centering
  \includegraphics[width=1.0\linewidth]{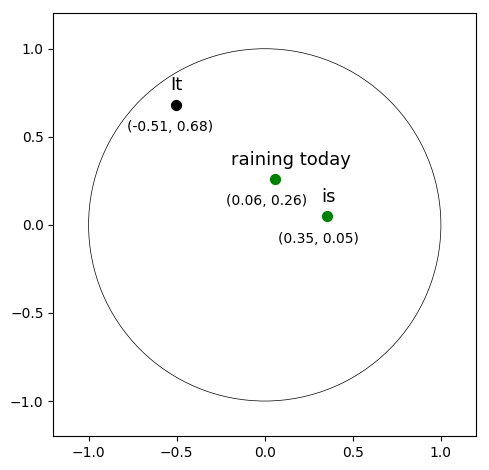}
  \caption{$\theta_{\text{is}} \oplus_M \theta_{\text{raining today}}$}
  \label{fig:mob_add_2}
\end{subfigure}%
\hspace{0.1\textwidth}
\begin{subfigure}{.45\textwidth}
  \centering
  \includegraphics[width=1.0\linewidth]{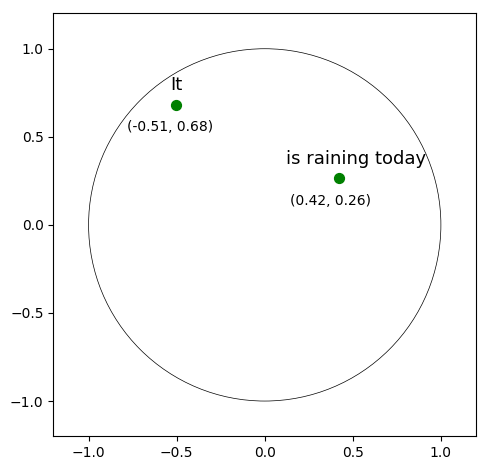}
  \caption{$\theta_{\text{It}} \oplus_M \theta_{\text{is raining today}}$}
  \label{ffig:mob_add_3}
\end{subfigure}

\begin{subfigure}{.45\textwidth}
  \centering
  \includegraphics[width=1.0\linewidth]{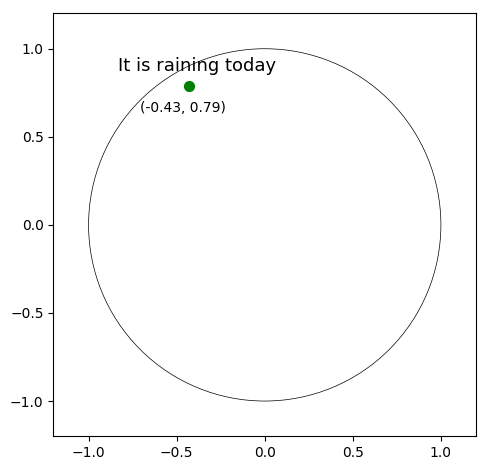}
  \caption{The final representation of the sentence example}
  \label{fig:mob_add_4}
\end{subfigure}%
\hspace{0.1\textwidth}

\caption{Sentence construction example}
\label{fig:mobius_summation_visualization}
\end{figure}

In Figure \ref{fig:mobius_summation_visualization} we can see a visualization of the aforementioned example in a 2-D Poincare disk.

\section{Our models}
Having defined the theoretical ground that we use to build upon, here we present the models that have been used. We first talk about a simpler model, presented in Section \ref{section:mobius_summation}, that doesn't use any additional trainable parameters, apart from the word embeddings. After that in Section \ref{section:mobius_summation_ffnn}, we introduce a model that combines the Poincare ball, as a space for the embeddings, together with a Feed Forward Neural Network for improving the classification performance.

\subsection{Mobius summation as a sentence representation} \label{section:mobius_summation}
For the first model we present, we only address Textual Entailment as a binary classification task, where samples are divided into two categories:
\begin{itemize}
	\item Positive: pairs that represent entailment
	\item Negative: pairs that do not represent entailment
\end{itemize} 
As a loss function, we use the max-margin loss, defined as:
\begin{equation}
	  \Loss = \sum_{(p,h) \in P}E(f(p), f(h)) + \sum_{(p',h') \in N} max\{0, \alpha - E(f(p'), f(h'))\}
\end{equation}

where $P$ represents the positive (entailment) samples, whereas $N$ represents the joint set of neutral and contradictive samples.$f(x)$ represents the sentence composition function as defined in Section \ref{section:sentence_representation} and $\alpha$ is margin classification hyperparameter. $E(u,v)$ represents the pair-score function which is defined as:

\begin{equation} \label{eq:energy_function}
	 E(u,v) = \beta d(u,v) + (1 - \beta) max\{0, \norm{v} - \norm{u}\}
\end{equation}

where $d(u,v)$ represents the Poincare distance function defined in Equation  (\ref{eq:poincare_distance}), $\beta$ represents a weighting hyperparameter between the distance between the respective sentences and their norm difference.

As we can see, the only trainable parameters in this model are the word embeddings themselves. The model solely relies on the poincare distance function and learning suitable word embeddings such that pairs that represent entailment have a small poincare distance, and pairs that do not are taken further apart. The second part of Equation (\ref{eq:energy_function}) favorizes sentences that are more abstract and entail other sentences to be closer to the origin.

The goal of this loss function is to make pairs that represent entailment have small scores, whereas pairs that don't represent entailment have big scores. As for evaluating and making predictions on unseen samples, a threshold on the validation set is picked after training the model, such that if a sample's pair-score function is below the threshold the sample is classified as entailment and otherwise it is not.

Given the nature of the loss function, this model is only applicable to solving the binary classification problem, where we merge the neutral and contradictive class into one class.
Although relatively simplistic, given that the model doesn't use any additional parameters apart from the word embeddings, it serves as a good basis for the model introduced in Section \ref{section:mobius_summation_ffnn}

\subsection{Mobius summation + FFNN} \label{section:mobius_summation_ffnn}
\begin{figure}[h!]
\begin{center}
	\begin{adjustwidth}{-1.9cm}{}
	\includegraphics[scale=0.5]{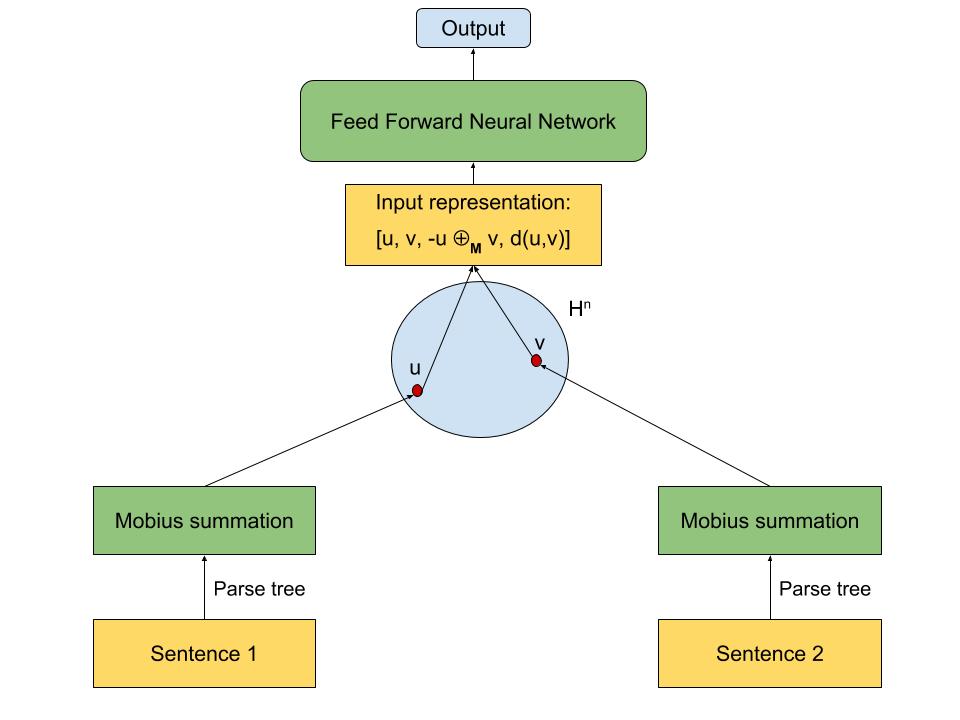}\\
	\end{adjustwidth}
	\caption{Model architecture for Mobius Summation with a Feed Forward Neural Network on top. The blue circle represents the Poincare ball as an embedding space for both individual word embeddings and composed sentences.}
	\label{fig:model_architecture}
\end{center}
\end{figure}

This model, in addition to having the same sentence composition method as the previously mentioned one, also has a Feed Forward Neural Network \cite{bebis1994feed} on top of the obtained sentence representations as shown in Figure \ref{fig:model_architecture}. This composition increases the complexity of the model and with it, among achieving better accuracy, enables us to solve the 3-way classification problem. The loss function is the cross-entropy loss function defined as:

\begin{equation}
	\Loss(x,y) = -\sum_{y_i}y_i \log y_i^{'}
\end{equation}

where $y$ represents a one-hot encoded vector for the ground truth labels and $y^{'}$ represents the soft-max values predicted by the model. The trainable parameters are the word embeddings, to which the same rules as in the previous model apply, and the weight parameters in the feed forward network. We take partial derivatives of the loss function with respect to every trainable parameter and update the parameters using the Stochastic Gradient Descent method which we talk about in Section \ref{section:Optimization}.
It is important to stress out that the parameters in the FFNN live in the euclidean space and are worked with and updated accordingly, without applying the Riemannian gradient on them.

\subsection{Mobius averaging + FFNN} \label{section:mobius_averaging_ffnn}
So far, we presented models that use recursive mobius summation as a method to represent sentences. Here we present the Mobius averaging model which in addition to using mobius summation for obtaining the sentence representation, it multiplies the obtained representation by the inverse of number of words in that sentence.
The idea for mobius averaging in the hyperbolic space comes as a natural counterpart to euclidean averaging.
Mobius scalar multiplication in a Poincare unit ball is defined in \cite{mobius} as:

\begin{equation} \label{eq:mobius_multiplication}
r \otimes_M v = \tanh\left(r\tanh^{-1}||v||\right)\frac{v}{||v||}
\end{equation}

Utilizing this definition, we could define a sentence representation as: 

\begin{equation}
\frac{1}{N} \otimes_M \left(w_1 \oplus_M\cdots\oplus_M w_N\right)
\end{equation}

where $N$ is the number of words in the sentence.\\

Some rules that apply to mobius multiplication are:

\begin{equation} \label{eq:mob_mult_rule_1}
	1 \otimes_M x = x
\end{equation}
\begin{equation} \label{eq:mob_mult_rule_2}
	(r_1 \cdot r_2) \otimes_M x = r_1 \otimes_M \left(r_2 \otimes_M x \right)\\
\end{equation}
\begin{equation} \label{eq:mob_mult_rule_3}
	(r_1 + r_2) \otimes_M x = \left(r_1 \otimes_M x \right) \oplus_M \left( r_2 \otimes_M x \right) 
\end{equation}

One of the motivating examples, for using mobius averaging as a sentence representation, is the fact that 
\begin{equation}
	x = \frac{1}{k}\otimes\left(\underbrace{x\oplus_M\cdots\oplus_M x}_{\text{k times}} \right) 
\end{equation}

\begin{proof}
In order to prove that the aforementioned equality holds we need to prove that 
\begin{equation}
	x = \frac{1}{k} \otimes \left( k \otimes x \right)
\end{equation}

\begin{equation}
	\frac{1}{k} \otimes \left( k \otimes x \right) = \left(\frac{1}{k}  k  \right) \otimes x = 1 \otimes x = x
\end{equation}

This holds because of the rules defined in Equations \ref{eq:mob_mult_rule_1} and \ref{eq:mob_mult_rule_2}.
	
Next, we need to prove that $k \otimes_M x = \left(\underbrace{x\oplus_M\cdots\oplus_M x}_{\text{k times}} \right)$ regardless of the parentheses order in the expression on the right side of the equality. 

It follows that:
\begin{equation}
	k \otimes_M x = \left(\underbrace{1 + \cdots + 1}_{\text{k times}} \right) \otimes_M x 
\end{equation}

If we substitute $r^{*} = \underbrace{1 + \cdots + 1}_{\text{k-1 times}}$, we get: 
\begin{equation}
\left(\underbrace{1 + \cdots + 1}_{\text{k times}} \right) \otimes_M x = \left(r^{*} + 1 \right) \otimes_M x = \left(r^{*} \otimes_M x  \right) \oplus_M \left(1 \otimes_M x \right)
\end{equation}

Then if we follow recursively, depending on the parenthesesition of $r^{*}$, we will end up with the following expression: 

\begin{equation}
\left(1 \otimes_M x \right) \oplus_M  \cdots  \oplus_M \left(1 \otimes_M x \right) = x \oplus_M \cdots \oplus_M x
\end{equation}

Which, if we decompose the $k$ into sum of ones with the same parenthesisasion  as the initial expression on the right-hand side (which is possible beause of associativity in addition), will lead us to the expression on the initial left hand side.
\end{proof}

\section{Spaces of constant curvature} \label{section:curvatures}
So far we have only examined the Poincare unit ball as an embedding space where every point $x$ has a norm less than one. In this section we present a c-hyperbolic space conformal with the Euclidean space defined as: $\left(\D_c, g^{c}\right)$, where $\D_c = \{ x \in \R^d : c\norm{x}^2 < 1\}$ and the Riemannian metric tensor is defined as: 

\begin{equation}
g_x^{\D} = \lambda_x^{2}g^{E}
\end{equation}

where $\lambda_x = \frac{2}{1 - c\norm{x}^2}$ and $g^{E}$ represents the Euclidean metric tensor of components \textbf{I}$_n$ of the standard embedding space $\R^d$.
The space is of constant curvature $c \in \R^{d}$. For $c = 0$ we obtain the Euclidean space, whereas for $c = 1$ we recover the Poincare unit ball that was discussed earlier. The operations defined in Equations \ref{eq:mobius_addition}, \ref{eq:mobius_multiplication}, \ref{eq:poincare_distance}, change to:

\begin{equation}
	u \oplus_M v = \frac{\left( 1 + 2 c \langle u , v \rangle + c \norm{v}^{2} \right)u + \left(1 - c  \norm{u}^2 \right)v}{1 + 2  c \langle u , v \rangle + c^{2}  \norm{u}^{2} \norm{v}^{2}}
\end{equation}

\begin{equation}
r \otimes_M v = \frac{1}{\sqrt{c}}\tanh\left(r\tanh^{-1}\sqrt{c}\norm{v} \right)\frac{v}{\norm{v}}, r \in \R, v \in \D_c
\end{equation}

\begin{equation}
	d(u,v) = \frac{2}{\sqrt{c}} \text{tanh}^{-1} \left( \sqrt{c} \norm{ -u \oplus_M v} \right)
\end{equation}

The goal of examining spaces of different constant curvature is to find an optimal embedding space for sentences which may not be a Poincare unit ball.
\section{Optimization}\label{section:Optimization}
Having defined the models that are being using in this thesis, we now turn to the Optimization part. The goal is to come up with the optimal word embeddings and FFNN weight parameters, such that the loss function defined in the previous sections is minimized. Formally, the objective is defined as:

\begin{equation}
\theta', W' \leftarrow \argmin_{\theta, W}\Loss(\theta, W)
\end{equation}

where $\theta$ represents the word embeddings and W represents the weight parameters in the Feed Forward Neural Network, if present.

We use Stochastic Gradient Descent \cite{bottou2010large} as an optimization method. For every sample in the training phase we take the partial derivatives of the loss function with respect to the parameters of the FFNN and the word embeddings that construct the respective sentences. As explained in Section \ref{section:riemannian_optimization}, for the word embeddings derivatives we apply  Riemannian gradient and the respective projection, if the updated embeddings get out of the unit ball. Since every sentence is represented as a recursive sum of mobius additions, the chain rule depends on the parsee tree construction of the sentence. This means, that for every sentence we might potentially have a different chain rule, depending of its constituency parse tree.

\subsection{Riemannian optimization in the poincare ball model} \label{section:riemannian_optimization}
Based on the word embeddings $\theta \in B^{d}$, we have a corresponding loss function  $\Loss(\theta)$. In order to minimize this loss function with respect to the word embeddings, we need to apply Riemannian Stochastic  Gradient Descent as defined in \cite{riemannian_optimization}. Parameters update using RSGD is defined as:

\begin{equation}
	\theta_{t+1} = R_{\theta_{t}}\left( -\eta_t \bigtriangledown_R\Loss(\theta_t) \right)
\end{equation}

where $R_{\theta_{t}}$ stands for retraction of $\theta$ onto $B$ at time $t$ and $\eta_t$ is the learning rate at time $t$.

In order for the points to stay within the Poincare ball we need to scale them if their norm is bigger than 1, using the following scaling rule:

\begin{equation}
    proj(\theta) =
               \begin{cases}
               \frac{\theta}{\norm{\theta} + \epsilon} \hspace{0.5cm} if \norm{\theta} \geq 1\\
               \theta\hspace{1.1cm} else
               \end{cases}
\end{equation}

or if we work with c-hyperbolic space as defined in Section \ref{section:curvatures}, the projection function is:
\begin{equation}
	proj(\theta) =
               \begin{cases}
               \theta \hspace{1.8cm} if  \hspace{0.2cm} c \norm{\theta}^{2} < 1 \\
               \frac{1}{\sqrt{c}}  \frac{\theta}{\norm{\theta} + \epsilon} \hspace{0.5cm} else
               \end{cases}                       
\end{equation}

where $\epsilon = 0.00001$ is used for rescaling the embeddings such that they don't lie on the ball boundary.

The update is done through taking the inverse of the Poincare ball metric tensor $g_\theta^{-1}$ which effectively leads us to the following update formula:

\begin{equation}
    \theta_{t + 1} \gets proj\left( \theta_t - \eta_t\frac{\left(1 - c\norm{\theta_t}^{2}\right)^{2}}{4} \bigtriangledown \Loss \right)
\end{equation}


%% file: datasets.tex


\chapter{Datasets}
\label{ch:datasets}
In this chapter we talk about the datasets that are being used throughout the experiments. Two of them are the standard and well known datasets for textual entailment: SNLI and SICK. The other two are synthetic datasets that we have constructed and suit as an illustrative proof about the learning process of the models presented in Chapter \ref{ch:core}.
\section{SNLI} \label{section:datasets_snli}
SNLI stands for Stanford Natural Language Inference and it has been introduced in \cite{snli_corpus}.
The SNLI corpus is a collection of 570k English sentence pairs manually labeled with labels of entailment, contradiction and neutral. \\

From Table \ref{table:dataset_snli} we can the number of pairs of sentences each of the datasets have. By its size the SNLI corpus is by far the largest corpus that addresses the textual entailment task  and is used in almost all of the approaches that try to tackle the problem of natural language inference.

\begin{center}
	\begin{tabular}{|c|c|}
	\hline
	\textbf{Training set size} & 550152\\
	\hline
	\textbf{Validation set size} & 10000\\
	\hline
	\textbf{Test set size} & 10000\\
	\hline
	\textbf{Vocabulary size} & 36983\\
	\hline
	\textbf{Entailment part} & 33.39\%\\
	\hline
	\textbf{Neutral part} & 33.27\%\\
	\hline
	\textbf{Contradiction part} & 33.34\%\\
	\hline
	\end{tabular}
	\captionof{table}{SNLI information}
	\label{table:dataset_snli}
\end{center}

The data was collected through Amazon Mechanical Turk. The Flickr30k corpus introduced in \cite{flickr30}, which is a dateset of images accompanied by their respective text descriptions,  was used to obtain the premises. After obtaining the premises, each worker was given a premise and a label (entailment, neutral and contradiction) and for each label the worker was asked to produce a hypothesis to the premise. This is important to stress out, because the dataset was constructed entirely by human effort and without any synthetic text creation. All together, 2500 workers were involved in producing the dataset.

A round of validation was performed on 10$\%$ of the data, to minimize the risk of data corruption and attest to corpus quality. Each of the validation sentence pairs was given to 4 different workers, without providing the gold label. So, in total, for each validation pair 5 labels were obtained. A gold label was assigned to a validation pair only if there were at least 3 same labels out of the 5 provided. Otherwise, the label of the pair was discarded, and although included in the SNLI dataset, the pair doesn't contribute to training or testing the models. Given its size, the corpus is very suitable and used for training models that require a large number of parameters.

\section{SICK} \label{section:datasets_sick}

The SICK dataset has been introduced in \cite{sick_corpus} and stands for Sentences Involving Compositional Knowledge. Unlike the SNLI dataset, SICK is much smaller and it only consists of 9.5K pairs of sentences. Nevertheless, it is used extensively in models that tackle the textual entailment problem. As we can see from Table \ref{table:dataset_sick} the split between training and test is roughly 50-50. Although this dataset is approximately 50 times smaller than the SNLI dataset it is useful for exploring how do models learn with much less data and whether they are able to quickly adapt. Unlike the SNLI dataset, which comes together with the sentences' parse trees, the SICK  consists of  raw sentences. In addition to the entailment relation label, each pair of sentences in the dataset has a sentence relatedness score assigned to it.

I used the stanford parser \cite{stanford_parser} to obtain the respective parse trees for the sentences that are used for the Mobius Summation models.

\begin{center}
	\begin{tabular}{|c|c|}
	\hline
	\textbf{Training set size} & 4500\\
	\hline
	\textbf{Test set size} & 4927\\
	\hline
	\textbf{Vocabulary size} & 2461\\
	\hline
	\textbf{Entailment part} & 28.87\%\\
	\hline
	\textbf{Neutral part} & 56.36\%\\
	\hline
	\textbf{Contradiction part} & 14.78\%\\
	\hline
	\end{tabular}
	\captionof{table}{SICK information}
	\label{table:dataset_sick}
\end{center}

\section{Adjective Noun Toy dataset} \label{section:datasets_adjnoun}
Here we present a dataset that was developed to test,analyze and emphasize some of the features the models that we use throughout the experiments have. The dataset is of a very simplistic nature, consisting of positive (entailment) and negative (non-entailment) samples. The vocabulary of size $N$ is evenly split between words that represent adjectives and words that represent nouns. For the sake of simplicity we consider the first $\frac{N}{2}$ of the words to be adjectives and the rest to be nouns.

The positive samples are pairs of nouns that entail those same nouns with an adjective as a qualifier. More formally, the positive samples are defined as:
\begin{center}

    $"noun_\textbf{i}"$ entails $"adjective_\textbf{j}$ $noun_\textbf{i}"$

\end{center}
where $1 \leq j \leq \frac{N}{2}$ and $ \frac{N}{2} < i \leq N$.\\

The negative samples on the other hand are defined as:
\begin{center}

    "$noun_\textbf{i}$" does not entail "$adjective_\textbf{j}$ $noun_\textbf{k}$"

\end{center}
where $1 \leq j \leq \frac{N}{2}$ and $ \frac{N}{2} < i,k \leq N$ and $i \neq k$.\\

\begin{center}
	\begin{tabular}{|c|c|}
	\hline
	\textbf{Training set size} & 2000\\
	\hline
	\textbf{Validation set size} & 20000\\
	\hline
	\textbf{Test set size} & 20000\\
	\hline
	\textbf{Vocabulary size} & 1000\\
	\hline
	\textbf{Entailment part} & 50.40\%\\
	\hline
	\textbf{Non-entailment part} & 49.60\%\\
	\hline
	\end{tabular}
	\captionof{table}{Adjective-noun dataset information}
	\label{table:dataset_adjnoun}
\end{center}

The dataset was constructed such that the training set has all of the nouns and adjectives appear at least once. The validation and test sets were made purposefully larger in order to assess the models' performance on bigger datasets, given relatively smaller number of training samples.
\section{Numbers toy dataset} \label{section:datasets_numbers}
Here we present the second dataset that we constructed in order to test and analyze the behavior of our baselines more thoroughly. The numbers dataset addresses the problem of entailment from a different perspective. We consider positive numbers that are smaller to entail those numbers that are bigger than themselves. The logic being is that smaller numbers are closer to the origin (in this case the object space is one-dimensional representing the positive side of the horizontal axis) and those that are bigger are further away. In other words, the goal is to learn the "less than" operation between numbers.

We restricted the dataset to only 4-digit numbers. As we can see in Table \ref{table:dataset_numbers} the dataset is divided into training, validation and test set. 
The dataset is evenly split between positive and negative samples.

\begin{center}
	\begin{tabular}{|c|c|}
	\hline
	\textbf{Numbers range} & 1000-9999\\
	\hline
	\textbf{Training set size} & 8000\\
	\hline
	\textbf{Validation set size} & 1000\\
	\hline
	\textbf{Test set size} & 1000\\
	\hline
	\textbf{Entailment part} & 50.06\%\\
	\hline
	\textbf{Non-entailment part} & 49.94\%\\
	\hline
	\end{tabular}
	\captionof{table}{Numbers dataset information}
	\label{table:dataset_numbers}
\end{center}

As a number representation method we present Left Mobius Summation as defined in Equation \ref{eq:left_mob_sum} and Right Mobius Summation defined in Equation \ref{eq:right_mob_sum}.

\begin{equation}\label{eq:left_mob_sum}
\text{LeftMobSum} (number) = d_1 \oplus_M (d_2 \oplus_M (d_3 \oplus_M d_4))
\end{equation}
\begin{equation}\label{eq:right_mob_sum}
\text{RightMobSum} (number) = ((d_1 \oplus_M d_2) \oplus_M d_3) \oplus_M d_4
\end{equation}

where $d_1$,  $d_2$,  $d_3$,  $d_4$ represent the digit embeddings for the first,second,third and fourth digit in the number, respectively.


%% file: experiments_chapter.tex



\chapter{Experiments}
\label{ch:experiments}

In this chapter we present the results obtained on the datasets introduced in Chapter \ref{ch:datasets}. We use methods introduced in Chapter \ref{ch:related_work} as baselines and the methods we discussed in Chapter \ref{ch:core}. For clarity, in Table \ref{table:methods_abbreviation} we present the abbreviations that we are going to use throughout this chapter. 

\begin{center} 
	\begin{adjustwidth}{-0.65cm}{}
	\begin{tabular}{|c|c|}
	\hline
	\textbf{Model name} & \textbf{Description}\\
	\hline
	\textbf{MS} & \makecell{Mobius summation using sentence parse tree \\ as was introduced in Section \ref{section:mobius_summation}} \\
	\hline
	\textbf{LMS} & \makecell{Mobius summation that sums the embeddings \\ from left to right.} \\
	\hline
	\textbf{RMS} & \makecell{Mobius summation that sums the embeddings \\ from right to left.} \\
	\hline
	\textbf{MS+FFNN} & \makecell{Mobius summation using sentence parse tree with FFNN on top \\ as introduced in Section \ref{section:mobius_summation_ffnn} }\\
	\hline
	\textbf{LMS+FFNN} & \makecell{Mobius summation that sums the embeddings \\ from left to right with FFNN on top.} \\
	\hline
	\textbf{RMS+FFNN} & \makecell{Mobius summation that sums the embeddings \\ from right to left with FFNN on top.} \\
	\hline
	\textbf{MS+FFNN$_{c = x}$} & \makecell{Mobius summation in a c-ball with FFNN on top \\ as introduced in Section \ref{section:curvatures} }\\
	\hline
	\textbf{MA+FFNN} & Mobius averaging as introduced in Section \ref{section:mobius_averaging_ffnn} \\
	\hline
	\textbf{EA+FFNN} & Euclidean Averaging as introduced in Section \ref{section:euclidean_averaging} \\
	\hline
	\textbf{LSTM+FFNN} & LSTMs model as introduced in Section \ref{section:lstm} \\
	\hline
	\textbf{OE} & Order embeddings as introduced in Section \ref{section:order_embeddings} \\
	
	\hline	
	\end{tabular}
	\end{adjustwidth}
	\captionof{table}{Methods abbreviation} 
	\label{table:methods_abbreviation}
\end{center}

We optimize  over different settings and outline the differences between models' performances. We divide the experiments along the embedding dimension and binary vs. three-way entailment classification. In the binary classification entailment represents one class and we merge the neutral and contradiction classes into one non-entailment class.
For the binary classification task, I am outlining the metrics:
\begin{equation}
	F_1 = 2 \cdot \frac{\text{precision} \cdot \text{recall}}{\text{precision} +\text{recall}}
\end{equation}
\begin{equation}
	\text{precision} =\frac{\text{true positives}}{\text{true positives} + \text{false positives}}
\end{equation}
\begin{equation}
	 \text{recall} = \frac{\text{true positives}}{\text{true positives} + \text{false negatives}}
\end{equation}
\begin{equation}
	 \text{accuracy} = \frac{\text{true positives} + \text{true negatives} + }{\text{true positives} + \text{false negatives} + \text{false positives} + \text{true negatives}}
\end{equation}

where

\begin{itemize}
	\item \textbf{true positives} represent pairs of entailment for which we also predicted entailment
	\item \textbf{false positives} represent pairs of non-entailment for which we  (wrongly) predicted entailment
	\item \textbf{true negatives} represent pairs of non-entailment for which we also predicted non-entailment
	\item \textbf{false negatives} represent pairs of entailment for which we (wrongly) predicted not-entailment
\end{itemize}

It is important to stress out that after each epoch we compute the scores \textbf{validation accuracy}, \textbf{test accuracy}, \textbf{f1}.
The triple which has the best validation accuracy is picked \footnote{We pick the best model judging by the validation because in real life scenario we wouldn't be able to actually have the test set and make decisions based on it.} and reported. In the three-way classification version, apart from the test and validation accuracy, we also report the confusion matrix where we present the observed vs. predicted values. 

The field \textbf{Concatenation method}, that is present in the tables of the following sections, represents the input representation which was fed to the Feed Forward Neural Network (where applicable) after obtaining the sentence representations for the premise and the hypothesis. For all the methods that use a Feed Forward Neural network we used a one-layer FFNN with 256 units and a softmax layer on top. 
For models \textbf{MS}, \textbf{LMS} and \textbf{RMS} we use a loss threshold $\alpha=0.05$ and a weight score parameter $\beta=0.5$.
We conducted all the experiments using the TensorFlow framework \cite{tensorflow}.

\section{Ablation studies} \label{section:experiments_ablation}
Here we present the various input representation we've tried on the SICK dataset using the \textbf{MS+FFNN} and  \textbf{MA+FFNN} methods. Getting the right set of features as an input to the Feed Forward Network is crucial to achieving good results further in the experiments. For the methods operating in the hyperbolic space it was very important the operations that were applied to the obtained sentence representations be defined in the hyperbolic space. For example, operations like $|u-v|, u*v$, which are present in the \textbf{EA+FFNN} and \textbf{LSTM+FFNN} methods, are not theoretically supported in the Poincare ball. 
\begin{center}  
	\begin{adjustwidth}{-0cm}{}
	\begin{tabular}{|c|c|c|c|c|}
	\hline
	\textbf{Model} & \textbf{\makecell{Concatenation \\ method}} & \textbf{Test accuracy} & \textbf{Test F1 score}\\
	\hline
	\textbf{\makecell{MA+FFNN}} & [$u,v,-u\oplus_Mv, \cos(u,v), d(u,v)$] & 83.95\% & 69.52\%\\
	\hline
	\textbf{\makecell{MA+FFNN}} & [$u,v,-u\oplus_Mv, d(u,v)$] & 82.40\% & 65.77\%\\
	\hline
	\textbf{\makecell{MA+FFNN}} & [$u,v$] & 71.14\% & 24.03\%\\
	\hline
	\textbf{\makecell{MA+FFNN}} & [$-u\oplus_Mv$] & 72.21\% & 12.75\%\\
	\hline
	\textbf{\makecell{MA+FFNN}} & [$u,v,-u\oplus_Mv$] & 71.91\% & 9.06\%\\
	\hline
	\textbf{\makecell{MS+FFNN}} & [$u,v,-u\oplus_Mv, \cos(u,v), d(u,v)$] & \textbf{86.77\%} & 76.71\%\\
	\hline
	\textbf{\makecell{MS+FFNN}} & [$u,v,-u\oplus_Mv, d(u,v)$] & \textbf{86.34\%} & 76.02\%\\
	\hline
	\textbf{\makecell{MS+FFNN}} & [$u,v$] & 71.12\% & 14.12\%\\
	\hline
	\textbf{\makecell{MS+FFNN}} & [$-u\oplus_Mv$] & 69.35\% & 6.32\%\\
	\hline
	\textbf{\makecell{MS+FFNN}} & [$u,v,-u\oplus_Mv$] & 77.94\% & 62.21\%\\
	\hline
	\end{tabular}
	\end{adjustwidth}
	\captionof{table}{SICK dataset 2 class, FFNN: 256, embedding dimension: 50, epochs: 70} 
	\label{table:abblation_50}
\end{center}

\begin{center} 
	\begin{adjustwidth}{0cm}{}
	\begin{tabular}{|c|c|c|c|c|}	
	\hline
	\textbf{Model} & \textbf{\makecell{Concatenation \\ method}} & \textbf{Test accuracy} & \textbf{Test F1 score}\\
	\hline
	\textbf{\makecell{MA+FFNN}} & [$u,v,-u\oplus_Mv, \cos(u,v), d(u,v)$] & 82.28\% & 65.80\%\\
	\hline
	\textbf{\makecell{MA+FFNN}} & $[u,v,-u\oplus_Mv, d(u,v)]$ & 82.32\% & 63.93\%\\
	\hline
	\textbf{\makecell{MA+FFNN}} & [$u,v$] & 71.08\% & 15.43\%\\
	\hline
	\textbf{\makecell{MA+FFNN}} & [$-u\oplus_Mv$] & 71.50\% & 11.59\%\\
	\hline
	\textbf{\makecell{MA+FFNN}} & [$u,v,-u\oplus_Mv$] & 72.09\% & 46.76\%\\
	\hline
	\textbf{\makecell{MS+FFNN}} & [$u,v,-u\oplus_Mv, \cos(u,v), d(u,v)$] & \textbf{85.31\%} & 75.49\%\\
	\hline
	\textbf{\makecell{MS+FFNN}} & $[u,v,-u\oplus_Mv, d(u,v)]$ & \textbf{85.16\%} & 73.42\%\\
	\hline
	\textbf{\makecell{MS+FFNN}} & [$u,v$] & 71.06\% & 21.99\%\\
	\hline
	\textbf{\makecell{MS+FFNN}} & [$-u\oplus_Mv$] & 69.64\% & 5.43\%\\
	\hline
	\textbf{\makecell{MS+FFNN}} & [$u,v,-u\oplus_Mv$] & 75.90\% & 58.36\%\\
	\hline
	
	\end{tabular}
	\end{adjustwidth}
	\captionof{table}{SICK dataset 2 class, FFNN: 256, embedding dimension: 5, epochs: 70} 
	\label{table:abblation_5}
\end{center}

As we can see from the results in Tables \ref{table:abblation_50} and \ref{table:abblation_5} the input representations [$u,v,-u\oplus_Mv, \cos(u,v), d(u,v)$] and $[u,v,-u\oplus_Mv, d(u,v)]$ yield consisently better results in both embedding dimension setups. This is expectable, due to the fact that the mentioned input representations bring more information to the FFNN in comparison to the rest of the concatenation methods. It is worth noting that the operation $\cos(u,v)$ is defined in the Poincare ball if we look at the points as points lying on the tangent space of the origin. So, if we consider $u$ and $v$ to be the respective projections from the points in the tangent space of the origin as defined in Equations \ref{eq:exp_map_u} and \ref{eq:exp_map_v}, then the equality defined in Equation \ref{eq:cos_equality} holds.
\begin{equation} \label{eq:exp_map_u}
	\widetilde{u} = exp\_map \left( u \right)
\end{equation}
\begin{equation}\label{eq:exp_map_v}
	\widetilde{v} = exp\_map \left( v \right)
\end{equation}
\begin{equation} \label{eq:cos_equality}
	\cos \left(u,v \right) = \cos \left( \widetilde{u}, \widetilde{v} \right) 
\end{equation}

\section{SNLI} \label{section:experiments_snli}
Here we present the results for the SNLI dataset that has been introduced in Section \ref{section:datasets_snli}. We evaluated across different hyperparameters for the models and report the best accuracies for each of the models, respectively. For the methods \textbf{MA+FFNN}, \textbf{MS+FFNN} we used a learning rate of 0.05 and for \textbf{LSTM+FFNN} we used a learning rate of 0.1. We used a learning rate of 0.001 for training the Order Embeddings. After trying various hidden state dimensions for LSTM, we report the best results that we got when using 128 as a hidden state dimension. For the Order Embeddings a hidden state of dimension 256 for the GRU unit was used as well. We also try different concatenation inputs and report only the best results we get for each model. We train all of the models until convergence.

\begin{center} 
	\begin{tabular}{|c|c|c|c|c|c|}
	\hline
	\textbf{Model} & \textbf{\makecell{Concatenation \\ method}} & \textbf{\makecell{Test \\ acc.}} & \textbf{\makecell{Test \\ F1}} \\
	\hline
	\textbf{\makecell{MS \\ epochs: 7}} & *  & 75.81\% & 60.76\% \\
	\hline
	\textbf{\makecell{MA+FFNN \\ epochs: 10}} & [$u,v,-u\oplus_Mv, \cos(u,v), d(u,v)$]  & 80.84\% & 72.40\% \\
	\hline
	\textbf{\makecell{MS+FFNN \\ epochs: 11}} & [$u,v,-u\oplus_Mv, \cos(u,v), d(u,v)$]  & 82.78\% & 75.89\% \\
	\hline
	\textbf{\makecell{MS+FFNN$_{c = 0.03}$ \\ epochs: 12}} & [$u,v,-u\oplus_Mv, \cos(u,v), d(u,v)$]  & 85.50\% & 78.85\% \\
	\hline
	\textbf{\makecell{EA+FFNN \\ epochs: 15}} & [$u,v,|u-v|, u*v, \langle u , v \rangle, euclid \textunderscore d(u,v)$] & 83.71\% & 76.07\% \\
	\hline
	\textbf{\makecell{LSTM+FFNN \\ epochs: 30}} & [$u,v,|u-v|, u*v$]  & 83.20\% & 75.82\% \\
	\hline
	\textbf{\makecell{OE \\ epochs: 15}} & * &  \textbf{88.29\%} & 81.98\% \\
	\hline
	\end{tabular}
	\captionof{table}{SNLI dataset 2 class, FFNN: 256, embedding dimension: 50} 
	\label{table:snli_results_report_embdim_50_class_2}
\end{center}

\begin{center} 
	\begin{tabular}{|c|c|c|c|c|c|}
	\hline
	\textbf{Model} & \textbf{\makecell{Concatenation \\ method}} & \textbf{\makecell{Test \\ acc.}} & \textbf{\makecell{Test \\ F1}} \\
	\hline
	\textbf{\makecell{MS \\ epochs: 7}} & * & 72.68\% & 53.35\% \\
	\hline
	\textbf{\makecell{MA+FFNN \\ epochs: 10}} & [$u,v,-u\oplus_Mv, \cos(u,v), d(u,v)$]  & 79.46\% & 69.79\% \\
	\hline
	\textbf{\makecell{MS+FFNN \\ epochs: 11}} & [$u,v,-u\oplus_Mv, d(u,v)$]  & 81.30\% & 73.21\% \\
	\hline
	\textbf{\makecell{MS+FFNN$_{c = 0.1}$ \\ epochs: 11}} & [$u,v,-u\oplus_Mv, d(u,v)$]  & 82.64\% & 74.18\% \\
	\hline
	\textbf{\makecell{EA+FFNN \\ epochs: 15}} & [$u,v,|u-v|, u*v, \langle u , v \rangle$]  & 81.45\% & 71.92\% \\
	\hline
	\textbf{\makecell{LSTM+FFNN \\ epochs: 30}} & [$u,v,|u-v|, u*v$]  & 81.93\% & 73.42\% \\
	\hline
	\textbf{\makecell{OE \\ epochs: 15}} & * &  \textbf{86.15\%} & 80.29\% \\
	\hline
	\end{tabular}
	\captionof{table}{SNLI dataset 2 class, FFNN: 256, embedding dimension: 5} 
	\label{table:snli_results_report_embdim_5_class_2}
\end{center}

As we can see in Tables  \ref{table:snli_results_report_embdim_50_class_2} and \ref{table:snli_results_report_embdim_5_class_2} the embedding dimension, apart from \textbf{LSTM+FFNN} model, does have an impact on the accuracy. This is not surprising since bigger embedding dimensionality, if given enough data, can be able to provide better embeddings. As expected from what was reported in \cite{order_embeddings}, the Order Embeddings are outperforming the other methods. The Mobius Summation with a space curvature, as was introduced in Section \ref{section:curvatures} is the second-best method. We observe that it achieves better results than Euclidean Averaging and Mobius Summation within a unit ball, giving us an insight that the optimal space for this family method may be something between a unit ball Hyperbolic space and a "flat" Euclidean space.

In the Tables \ref{table:snli_results_report_embdim_50_class_3} and \ref{table:snli_results_report_embdim_5_class_3} we can see the result for the three class problem, for embedding dimensions 50 and 5, respectively. We do not include \textbf{OE} and \textbf{MS} methods, since they are limited to only solving the binary classifcation entailment problem.

\begin{center} 
	\begin{tabular}{|c|c|c|c|c|c|}
	\hline
	\textbf{Model} & \textbf{\makecell{Concatenation \\ method}} & \textbf{\makecell{Test \\ acc.}}\\
	\hline
	\textbf{\makecell{MA+FFNN \\ epochs: 10}} & [$u,v,-u\oplus_Mv, \cos(u,v), d(u,v)$]  & 70.22\% \\
	\hline
	\textbf{\makecell{MS+FFNN \\ epochs: 10}} & [$u,v,-u\oplus_Mv, d(u,v)$] & 73.62\% \\
	\hline
	\textbf{\makecell{MS+FFNN$_{c = 0.01}$ \\ epochs: 10}} & [$u,v,-u\oplus_Mv, d(u,v)$]  & \textbf{76.82\%}\\
	\hline
	\textbf{\makecell{EA+FFNN \\ epochs: 15}} & [$u,v,|u-v|, u*v, \langle u , v \rangle, euclid \textunderscore d(u,v)$] & 75.22\%  \\
	\hline
	\textbf{\makecell{LSTM+FFNN \\ epochs: 30}} & [$u,v,|u-v|, u*v$] & 75.18\%  \\
	\hline
	
	\end{tabular}
	\captionof{table}{SNLI dataset 3 class, FFNN: 256, embedding dimension: 50} 
	\label{table:snli_results_report_embdim_50_class_3}
\end{center}

\begin{center} 
	\begin{tabular}{|c|c|c|c|c|c|}
	\hline
	\textbf{Model} & \textbf{\makecell{Concatenation \\ method}} & \textbf{\makecell{Test \\ acc.}} \\
	\hline
	\textbf{\makecell{MA+FFNN \\ epochs: 10}} & [$u,v,-u\oplus_Mv, \cos(u,v), d(u,v)$]  & 68.83\% \\
	\hline
	\textbf{\makecell{MS+FFNN \\ epochs: 11}} & [$u,v,-u\oplus_Mv, \cos(u,v), d(u,v)$]  & 71.31\%\\
	\hline
	\textbf{\makecell{MS+FFNN$_{c = 0.1}$ \\ epochs: 10}} & [$u,v,-u\oplus_Mv, d(u,v)$] & \textbf{72.94\%}\\
	\hline
	\textbf{\makecell{EA+FFNN \\ epochs: 15}} & [$u,v,|u-v|, u*v$]  & 71.36\% \\
	\hline
	\textbf{\makecell{LSTM+FFNN \\ epochs: 30}} &  [$u,v,|u-v|, u*v$]   & 71.85\%  \\
	\hline
	
	\end{tabular}
	\captionof{table}{SNLI dataset 3 class, FFNN: 256, embedding dimension: 5} 
	\label{table:snli_results_report_embdim_5_class_3}
\end{center}

\setlength\unitlength{1.5cm}
\begin{figure*}[t!]
\centering
\begin{subfigure}[t]{.45\textwidth}
\begin{picture}(4,4)
\multiput(0.1,0.1)(0,1){4}{\line(1,0){3}}
\multiput(0.1,0.1)(1,0){4}{\line(0,1){3}}
\put(0.5,2.5){2432}
\put(1.5,2.5){312}
\put(2.5,2.5){492}

\put(0.5,1.5){485}
\put(1.5,1.5){1803}
\put(2.5,1.5){931}

\put(0.5,0.5){233}
\put(1.5,0.5){294}
\put(2.5,0.5){2841}

\put(-0.4,2.5){C}
\put(-0.4,1.5){N}
\put(-0.4,0.5){E}

\put(0.5,3.3){C}
\put(1.5,3.3){N}
\put(2.5,3.3){E}

\put (-0.85, 1.0){\rotatebox{90}{\textbf{Observed}}}
\put(1.0,3.75){\textbf{Predicted}}
\end{picture}
\caption{\textbf{MA+FFNN} method}
\end{subfigure}\hfill%
~
\begin{subfigure}[t]{.45\textwidth}

\begin{picture}(4,4) 
\multiput(0.1,0.1)(0,1){4}{\line(1,0){3}}
\multiput(0.1,0.1)(1,0){4}{\line(0,1){3}}
\put(0.5,2.5){2378}
\put(1.5,2.5){436}
\put(2.5,2.5){422}

\put(0.5,1.5){465}
\put(1.5,1.5){2104}
\put(2.5,1.5){650}

\put(0.5,0.5){221}
\put(1.5,0.5){397}
\put(2.5,0.5){2750}

\put(-0.4,2.5){C}
\put(-0.4,1.5){N}
\put(-0.4,0.5){E}

\put(0.5,3.3){C}
\put(1.5,3.3){N}
\put(2.5,3.3){E}

\put (-0.85,1.0){\rotatebox{90}{\textbf{Observed}}}
\put(1.0,3.75){\textbf{Predicted}}
\end{picture}
\caption{\textbf{MS+FFNN} method}
\end{subfigure}
~
\begin{subfigure}[t]{.45\textwidth}
\begin{picture}(4,4)
\multiput(0.1,0.1)(0,1){4}{\line(1,0){3}}
\multiput(0.1,0.1)(1,0){4}{\line(0,1){3}}
\put(0.5,2.5){2383}
\put(1.5,2.5){484}
\put(2.5,2.5){370}

\put(0.5,1.5){389}
\put(1.5,1.5){2323}
\put(2.5,1.5){507}

\put(0.5,0.5){228}
\put(1.5,0.5){481}
\put(2.5,0.5){2659}

\put(-0.4,2.5){C}
\put(-0.4,1.5){N}
\put(-0.4,0.5){E}

\put(0.5,3.3){C}
\put(1.5,3.3){N}
\put(2.5,3.3){E}

\put (-0.85, 1.0){\rotatebox{90}{\textbf{Observed}}}
\put(1.0,3.75){\textbf{Predicted}}
\end{picture}
\caption{\textbf{LSTM+FFNN} method}
\end{subfigure}\hfill%
~
\begin{subfigure}[t]{.45\textwidth}

\begin{picture}(4,4)
 
\multiput(0.1,0.1)(0,1){4}{\line(1,0){3}}
\multiput(0.1,0.1)(1,0){4}{\line(0,1){3}}
\put(0.5,2.5){2506}
\put(1.5,2.5){499}
\put(2.5,2.5){231}

\put(0.5,1.5){427}
\put(1.5,1.5){2324}
\put(2.5,1.5){468}

\put(0.5,0.5){276}
\put(1.5,0.5){533}
\put(2.5,0.5){2559}

\put(-0.4,2.5){C}
\put(-0.4,1.5){N}
\put(-0.4,0.5){E}

\put(0.5,3.3){C}
\put(1.5,3.3){N}
\put(2.5,3.3){E}

\put (-0.85,1.0){\rotatebox{90}{\textbf{Observed}}}
\put(1.0,3.75){\textbf{Predicted}}
\end{picture}
\caption{\textbf{EA+FFNN} method}
\end{subfigure}
~
\begin{subfigure}[t]{1.0\textwidth}
\begin{picture}(4,4)
\multiput(0.1,0.1)(0,1){4}{\line(1,0){3}}
\multiput(0.1,0.1)(1,0){4}{\line(0,1){3}}	
\put(0.5,2.5){2462}
\put(1.5,2.5){496}
\put(2.5,2.5){278}

\put(0.5,1.5){358}
\put(1.5,1.5){2334}
\put(2.5,1.5){527}

\put(0.5,0.5){167}
\put(1.5,0.5){451}
\put(2.5,0.5){2750}

\put(-0.4,2.5){C}
\put(-0.4,1.5){N}
\put(-0.4,0.5){E}

\put(0.5,3.3){C}
\put(1.5,3.3){N}
\put(2.5,3.3){E}

\put (-0.85, 1.0){\rotatebox{90}{\textbf{Observed}}}
\put(1.0,3.75){\textbf{Predicted}}
\end{picture}
\caption{\textbf{MS+FFNN$_{c=0.01}$} method}
\end{subfigure}
\caption{Confusion matrices for methods \textbf{MA+FFNN}, \textbf{MS+FFNN}, \textbf{LSTM+FFNN}, \textbf{EA+FFNN} with embedding dimension 50 on the SNLI dataset for the three-way entailment problem. \textbf{C} stands for the class Contradiction, \textbf{N} stands for Neutral and \textbf{E} stands for Entailment.}
\label{fig:confusion_matrix_snli}
\end{figure*}

As we can see from Tables \ref{table:snli_results_report_embdim_50_class_3} and \ref{table:snli_results_report_embdim_5_class_3} there is a significant drop in accuracy for the \textbf{MS+FFNN} model when trying to solve the three-way entailment problem. The intuitive explanation is that the hyperbolic spaces may be suitable for detecting whether one object is entailed by another one, but they may not be as good for more subtle differentiations between the neutral and condraction class. In Figure \ref{fig:confusion_matrix_snli} we present the confusion matrices which tells us that the \textbf{MA+FFNN}, \textbf{MS+FFNN}, \textbf{MS+FFNN$_{c = 0.01}$} models are significantly better than the rest of the baselines when trying to predict entailment.
\begin{figure}
\begin{subfigure}{.45\textwidth}
  \centering
  \includegraphics[width=1.0\linewidth]{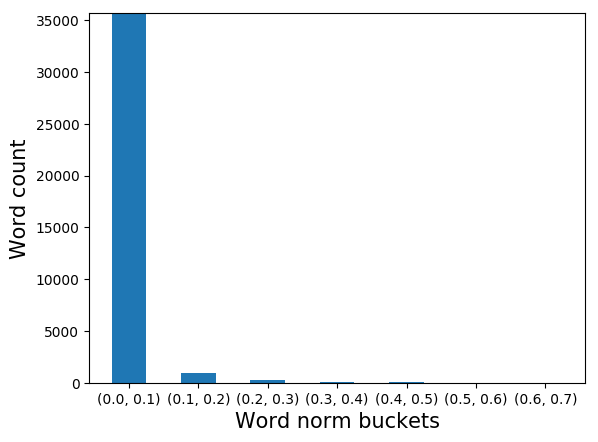}
  \caption{Epoch 1}
  \label{subfig:norms_snli_ms_epoch_0}
\end{subfigure}%
\hspace{0.1\textwidth}
\begin{subfigure}{.45\textwidth}
  \centering
  \includegraphics[width=1.0\linewidth]{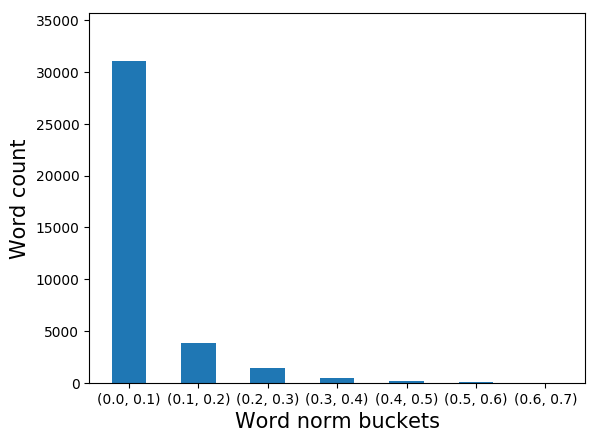}
  \caption{Epoch 4}
  \label{subfig:norms_snli_ms_epoch_5}
\end{subfigure}

\begin{subfigure}{.45\textwidth}
  \centering
  \includegraphics[width=1.0\linewidth]{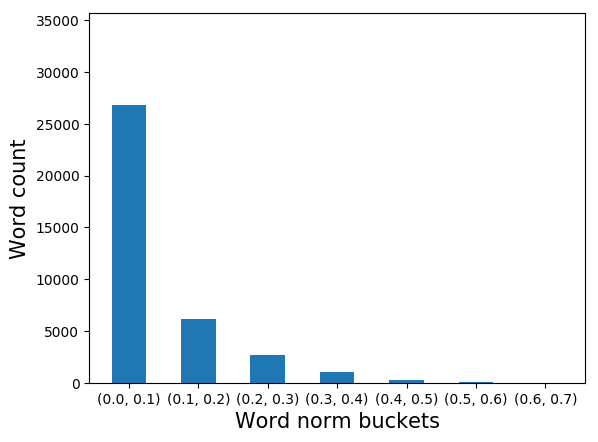}
  \caption{Epoch 7}
  \label{subfig:norms_snli_ms_epoch_10}
  
\end{subfigure}%
\hspace{0.1\textwidth}
\begin{subfigure}{.45\textwidth}
  \centering
  \includegraphics[width=1.0\linewidth]{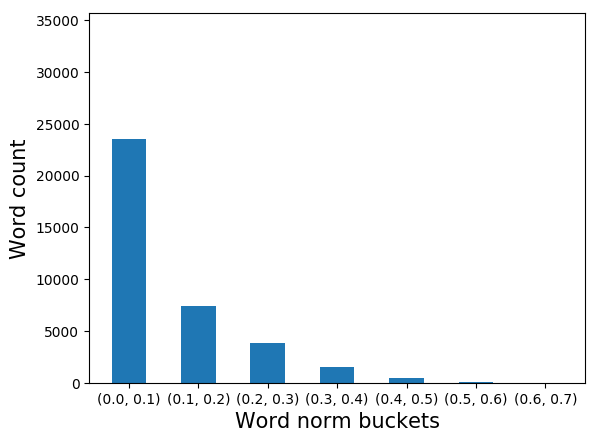}
  \caption{Epoch 10}
  \label{subfig:norms_snli_ms_epoch_15}
\end{subfigure}

\caption{Word embeddings norm distribution for method \textbf{MS+FFNN} for the binary classification problem with an embedding dimension 50}
\label{fig:norms_snli_ms}
\end{figure}

\begin{figure}
\begin{subfigure}{.45\textwidth}
  \centering
  \includegraphics[width=1.0\linewidth]{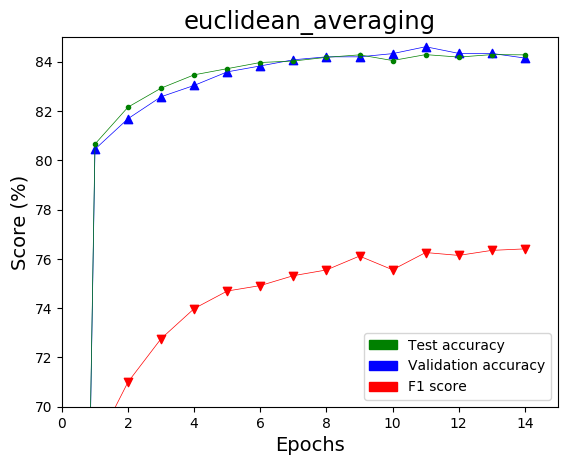}
  \caption{Euclidean averaging}
  \label{fig:euclid_avg}
\end{subfigure}%
\hspace{0.1\textwidth}
\begin{subfigure}{.45\textwidth}
  \centering
  \includegraphics[width=1.0\linewidth]{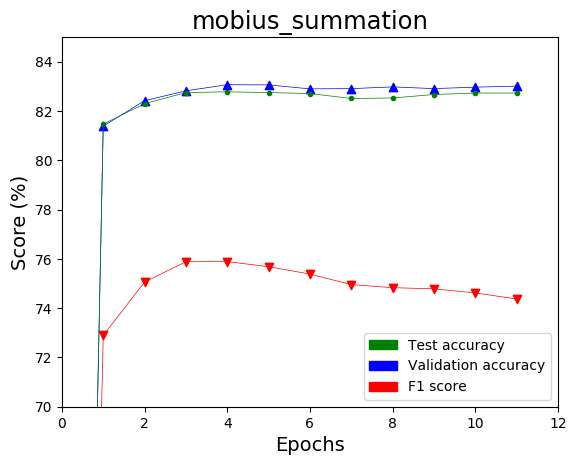}
  \caption{Mobius summation (unit ball)}
  \label{fig:mob_sum}
\end{subfigure}

\begin{subfigure}{.45\textwidth}
  \centering
  \includegraphics[width=1.0\linewidth]{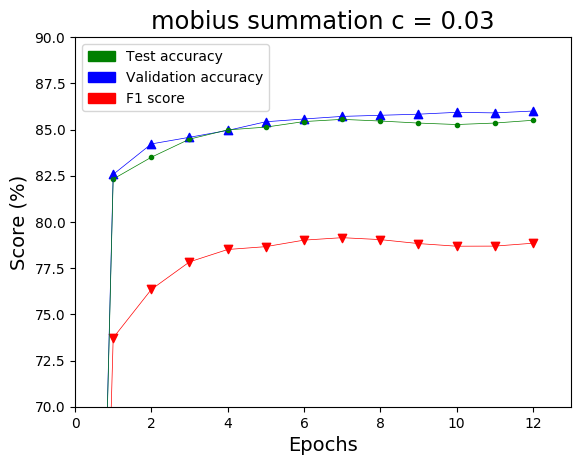}
  \caption{Mobius summation with c = 0.03}
  \label{fig:mob_sum_c}
  
\end{subfigure}%
\hspace{0.1\textwidth}
\begin{subfigure}{.45\textwidth}
  \centering
  \includegraphics[width=1.0\linewidth]{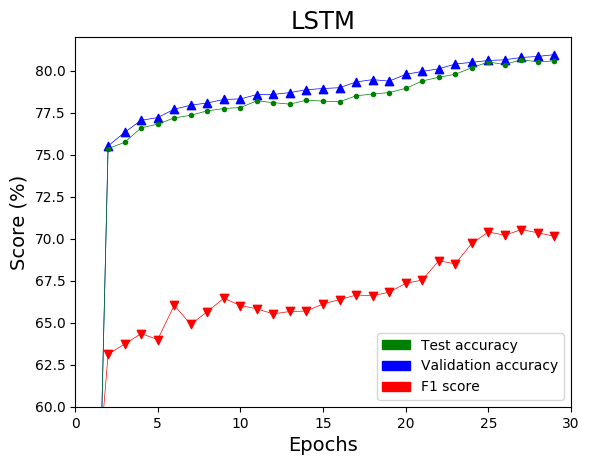}
  \caption{LSTM learning curve}
  \label{ffig:lc_lstm}
\end{subfigure}
\caption{Learning curves for some of the models introduced in Table \ref{table:snli_results_report_embdim_50_class_2}}
\label{fig:learning_curves}
\end{figure}

From Figure \ref{fig:learning_curves} we can see that the Mobius Summation method converges fairly quickly. In Figure \ref{fig:norms_snli_ms} we present a visualization on how embeddings' norm distribution changes over time for the \textbf{MS+FFNN} method.

\section{SICK}\label{section:experiments_sick}
Here we present the results for the SICK dataset that has been introduced in Section \ref{section:datasets_sick}. Since the SICK dataset is much smaller in comparison to SNLI, we trained the models for comparatively more epochs. 
It is important to note that the \textbf{OE} method requires a validation dataset for  coming up with a threshold that it uses for classifying entailment vs. non-entailment. Since the SICK dataset is split only between a training and a test set, we purposefully gave an unfair advantage of the \textbf{OE} method such that we compute the classification threshold and evaluate on the same test dataset. For \textbf{OE}, the GRU unit dimensionality is 512.

\begin{center} 
	\begin{tabular}{|c|c|c|c|c|c|c|}
	\hline
	\textbf{Model} & \textbf{\makecell{Concatenation \\ method}} & \textbf{\makecell{Test \\ acc.}} & \textbf{\makecell{Test \\ F1}} \\
	\hline
	\textbf{\makecell{MA+FFNN \\ epochs: 70}} & [$u,v,-u\oplus_Mv, \cos(u,v), d(u,v)$]  & 85.59\% & 75.24\% \\
	\hline
	\textbf{\makecell{MS+FFNN \\ epochs: 70}} & [$u,v,-u\oplus_Mv, \cos(u,v), d(u,v)$] & \textbf{86.77\%} & 76.71\% \\
	\hline
	\textbf{\makecell{MS+FFNN$_{c = 0.1}$ \\ epochs: 70}} & [$u,v,-u\oplus_Mv, d(u,v)$] & 86.65\% & 76.21\%  \\
	\hline
	\textbf{\makecell{EA+FFNN \\ epochs: 100}} & [$u,v,|u-v|, u*v, \langle u , v \rangle, euclid \textunderscore d(u,v)$] & 85.55\% & 73.76\% \\
	\hline
	\textbf{\makecell{LSTM+FFNN \\ epochs: 150}} & [$u,v,|u-v|, u*v$] & 75.50\% & 54.50\% \\
	\hline
	\textbf{\makecell{OE \\ epochs: 70}} & * & 85.22\% & 76.71\% \\
	\hline
	\end{tabular}
	\captionof{table}{SICK dataset 2 class, FFNN: 256, embedding dimension: 50} 
	\label{table:sick_results_report_embdim_50_class_2}
\end{center}

\begin{center} 
	\begin{tabular}{|c|c|c|c|c|c|c|}
	\hline
	\textbf{Model} & \textbf{\makecell{Concatenation \\ method}} & \textbf{\makecell{Test \\ acc.}} & \textbf{\makecell{Test \\ F1}} \\
	\hline
	\textbf{\makecell{MA+FFNN \\ epochs: 70}} & [$u,v,-u\oplus_Mv, \cos(u,v), d(u,v)$]  & 83.95\% & 70.00\% \\
	\hline
	\textbf{\makecell{MS+FFNN \\ epochs: 70}} & [$u,v,-u\oplus_Mv, d(u,v)$] & 85.31\% & 75.49\% \\
	\hline
	\textbf{\makecell{MS+FFNN$_{c = 0.1}$ \\ epochs: 70}} & [$u,v,-u\oplus_Mv, d(u,v)$] & \textbf{86.28\%} & 74.96\%  \\
	\hline
	\textbf{\makecell{EA+FFNN \\ epochs: 70}} & [$u,v,|u-v|, u*v, \langle u , v \rangle$] &  83.15\% & 70.46\% \\
	\hline
	\textbf{\makecell{LSTM+FFNN \\ epochs: 200}} & [$u,v,|u-v|, u*v$] & 75.32\% & 50.89\% \\
	\hline
	\textbf{\makecell{OE \\ epochs: 70}} & * & 83.96\% & 75.91\% \\
	\hline
	\end{tabular}
	\captionof{table}{SICK dataset 2 class, FFNN: 256, embedding dimension: 5} 
	\label{table:sick_results_report_embdim_5_class_2}
\end{center}
As we can see from Tables \ref{table:sick_results_report_embdim_50_class_2} and \ref{table:sick_results_report_embdim_5_class_2} the methods \textbf{MS+FFNN} and \textbf{MS+FFNN$_{c=0.1}$} are superior to the other methods, both in terms of accuracy and $F1$ score. For the \textbf{LSTM+FFNN} method I tried various LSTM hidden state sizes: 1024,512,256,128 with different learning rates: 0.5, 0.1, 0.05 and different input representations of $[u,v,|u-v|]$, $[u,v,|u-v|, \langle u, v \rangle]$ and $[u,v,|u-v|, \langle u, v \rangle, euclid\_dist(u,v)]$ and reported the best results, respectively. However, we can see a significant relative dropout in performance in comparison to the results from Section \ref{section:experiments_snli}. This is due to the fact that the SICK dataset is much smaller than SNLI and with it may not be as suitable for models consisting of more trainable parameters like \textbf{LSTM+FFNN}, as SNLI is.

Analogously to Section \ref{section:experiments_snli} we also present the results for the 3-way classification problem in Tables \ref{table:sick_results_report_embdim_50_class_3} and \ref{table:sick_results_report_embdim_5_class_3}.

\begin{center} 
	\begin{tabular}{|c|c|c|c|c|}
	\hline
	\textbf{Model} & \textbf{\makecell{Concatenation \\ method}} & \textbf{\makecell{Test \\ acc.}} \\
	\hline
	\textbf{\makecell{MA+FFNN \\ epochs: 10}} & [$u,v,-u\oplus_Mv, \cos(u,v), d(u,v)$]  & 71.28\% \\
	\hline
	\textbf{\makecell{MS+FFNN \\ epochs: 70}} & [$u,v,-u\oplus_Mv, \cos(u,v), d(u,v)$] & 76.72\% \\
	\hline
	\textbf{\makecell{MS+FFNN$_{c = 0.001}$ \\ epochs: 70}} & [$u,v,|-u\oplus_Mv|$] & \textbf{81.17\%}  \\
	\hline
	\textbf{\makecell{EA+FFNN \\ epochs: 70}} & [$u,v,|u-v|, u*v, \langle u , v \rangle, euclid \textunderscore d(u,v)$] & 80.62\% \\
	\hline
	\textbf{\makecell{LSTM+FFNN \\ epochs: 200}} & [$u,v,|u-v|, u*v, \langle u , v \rangle, euclid \_ d(u,v)$] & 67.95\% \\
	\hline
	\end{tabular}
	\captionof{table}{SICK dataset 3 class, FFNN: 256, embedding dimension: 50} 
	\label{table:sick_results_report_embdim_50_class_3}
\end{center}

\begin{center} 
	\begin{tabular}{|c|c|c|c|c}
	\hline
	\textbf{Model} & \textbf{\makecell{Concatenation \\ method}} & \textbf{\makecell{Test \\ acc.}}\\
	\hline
	\textbf{\makecell{MA+FFNN \\ epochs: 70}} & [$u,v,-u\oplus_Mv, \cos(u,v), d(u,v)$]  & 70.27\% \\
	\hline
	\textbf{\makecell{MS+FFNN \\ epochs: 70}} & [$u,v,-u\oplus_Mv, d(u,v)$] & 70.81\% \\
	\hline
	\textbf{\makecell{MS+FFNN$_{c = 0.0001}$ \\ epochs: 70}} & [$u,v,|-u\oplus_Mv|$] & \textbf{79.89\%}  \\
	\hline
	\textbf{\makecell{EA+FFNN \\ epochs: 70}} & [$u,v,|u-v|, u*v, \langle u , v \rangle, euclid \_ d(u,v)$] &  78.65\% \\
	\hline
	\textbf{\makecell{LSTM+FFNN \\ epochs: 150}} & [$u,v,|u-v|, u*v$] &  66.49\% \\
	\hline
	\end{tabular}
	\captionof{table}{SICK dataset 3 class, FFNN: 256, embedding dimension: 5} 
	\label{table:sick_results_report_embdim_5_class_3}
\end{center}

\setlength\unitlength{1.5cm}
\begin{figure*}[t!]
\centering
\begin{subfigure}[t]{.45\textwidth}
\begin{picture}(4,4)
\multiput(0.1,0.1)(0,1){4}{\line(1,0){3}}
\multiput(0.1,0.1)(1,0){4}{\line(0,1){3}}
\put(0.5,2.5){429}
\put(1.5,2.5){254}
\put(2.5,2.5){37}

\put(0.5,1.5){165}
\put(1.5,1.5){2296}
\put(2.5,1.5){332}

\put(0.5,0.5){48}
\put(1.5,0.5){453}
\put(2.5,0.5){913}

\put(-0.4,2.5){C}
\put(-0.4,1.5){N}
\put(-0.4,0.5){E}

\put(0.5,3.3){C}
\put(1.5,3.3){N}
\put(2.5,3.3){E}

\put (-0.85, 1.0){\rotatebox{90}{\textbf{Observed}}}
\put(1.0,3.75){\textbf{Predicted}}
\end{picture}
\caption{\textbf{MA+FFNN} method}
\end{subfigure}\hfill%
~
\begin{subfigure}[t]{.45\textwidth}

\begin{picture}(4,4) 
\multiput(0.1,0.1)(0,1){4}{\line(1,0){3}}
\multiput(0.1,0.1)(1,0){4}{\line(0,1){3}}
\put(0.5,2.5){505}
\put(1.5,2.5){182}
\put(2.5,2.5){33}

\put(0.5,1.5){217}
\put(1.5,1.5){2191}
\put(2.5,1.5){385}

\put(0.5,0.5){28}
\put(1.5,0.5){302}
\put(2.5,0.5){1084}

\put(-0.4,2.5){C}
\put(-0.4,1.5){N}
\put(-0.4,0.5){E}

\put(0.5,3.3){C}
\put(1.5,3.3){N}
\put(2.5,3.3){E}

\put (-0.85,1.0){\rotatebox{90}{\textbf{Observed}}}
\put(1.0,3.75){\textbf{Predicted}}
\end{picture}
\caption{\textbf{MS+FFNN} method}
\end{subfigure}
~
\begin{subfigure}[t]{.45\textwidth}
\begin{picture}(4,4)
\multiput(0.1,0.1)(0,1){4}{\line(1,0){3}}
\multiput(0.1,0.1)(1,0){4}{\line(0,1){3}}	
\put(0.5,2.5){463}
\put(1.5,2.5){223}
\put(2.5,2.5){34}

\put(0.5,1.5){138}
\put(1.5,1.5){2065}
\put(2.5,1.5){590}

\put(0.5,0.5){9}
\put(1.5,0.5){585}
\put(2.5,0.5){820}

\put(-0.4,2.5){C}
\put(-0.4,1.5){N}
\put(-0.4,0.5){E}

\put(0.5,3.3){C}
\put(1.5,3.3){N}
\put(2.5,3.3){E}

\put (-0.85, 1.0){\rotatebox{90}{\textbf{Observed}}}
\put(1.0,3.75){\textbf{Predicted}}
\end{picture}
\caption{\textbf{LSTM+FFNN} method}
\end{subfigure}\hfill%
~
\begin{subfigure}[t]{.45\textwidth}

\begin{picture}(4,4)
 
\multiput(0.1,0.1)(0,1){4}{\line(1,0){3}}
\multiput(0.1,0.1)(1,0){4}{\line(0,1){3}}
\put(0.5,2.5){552}
\put(1.5,2.5){148} 
\put(2.5,2.5){20} 

\put(0.5,1.5){84} 
\put(1.5,1.5){2407} 
\put(2.5,1.5){302} 

\put(0.5,0.5){12} 
\put(1.5,0.5){389} 
\put(2.5,0.5){1013} 

\put(-0.4,2.5){C}
\put(-0.4,1.5){N}
\put(-0.4,0.5){E}

\put(0.5,3.3){C}
\put(1.5,3.3){N}
\put(2.5,3.3){E}

\put (-0.85,1.0){\rotatebox{90}{\textbf{Observed}}}
\put(1.0,3.75){\textbf{Predicted}}
\end{picture}
\caption{\textbf{EA+FFNN} method}
\end{subfigure}
~
\begin{subfigure}[t]{1.0\textwidth}
\begin{picture}(4,4)
\multiput(0.1,0.1)(0,1){4}{\line(1,0){3}}
\multiput(0.1,0.1)(1,0){4}{\line(0,1){3}}
\put(0.5,2.5){564}
\put(1.5,2.5){139}
\put(2.5,2.5){17}

\put(0.5,1.5){80}
\put(1.5,1.5){2418}
\put(2.5,1.5){295}

\put(0.5,0.5){4}
\put(1.5,0.5){393}
\put(2.5,0.5){1017}

\put(-0.4,2.5){C}
\put(-0.4,1.5){N}
\put(-0.4,0.5){E}

\put(0.5,3.3){C}
\put(1.5,3.3){N}
\put(2.5,3.3){E}

\put (-0.85, 1.0){\rotatebox{90}{\textbf{Observed}}}
\put(1.0,3.75){\textbf{Predicted}}
\end{picture}
\caption{\textbf{MS+FFNN$_{c=0.001}$} method}
\end{subfigure}
\caption{Confusion matrices for methods \textbf{MA+FFNN}, \textbf{MS+FFNN}, \textbf{LSTM+FFNN}, \textbf{EA+FFNN} with embedding dimension 50 on the SICK dataset for the three-way entailment problem. \textbf{C} stands for the class Contradiction, \textbf{N} stands for Neutral and \textbf{E} stands for Entailment.}
\label{fig:confusion_matrix_sick}
\end{figure*}

We can see a clear difference of comparative performances between models relative to the results obtained on the SNLI Dataset in Section \ref{section:experiments_snli}. One obvious explanation is the dataset size and with SNLI being two orders of magnitude larger than SICK, one could argue that it has more data to train more complex models such as \textbf{OE}. Another reason is that, as shown in Sections \ref{section:datasets_snli} and \ref{section:datasets_sick}, the SICK dataset class distribution is heavily skewed towards samples of the neutral class. SNLI on the other hand, has a relatively equal distribution between the three classes. Even though the \textbf{MS+FFNN} method beats Order Embeddings on the SICK dataset, it is the abovementioned reason and the dataset size that lead me to conclude that the results obtained on the SNLI are more relevant.
In Figure \ref{fig:confusion_matrix_sick} we can see the confusion matrices produced by the respective models.

\section{4-digit numbers}\label{section:experiments_numbers}
Here we present the results for the 4-digit numbers toy dataset that has been introduced in Section \ref{section:datasets_numbers}. Since we do not work with sentences in this dataset, we use left and right mobius summation as defined in Equations \ref{eq:left_mob_sum} and \ref{eq:right_mob_sum}, respectively.

As we can see from Tables \ref{table:numbers_results_embdim_50} and  \ref{table:numbers_results_embdim_5} there isn't much of a difference whether we parenthesize the number representation from the left or from the right.

\begin{center} 
	\begin{tabular}{|c|c|c|c|c|c|}
	\hline
	\textbf{Model} & \textbf{\makecell{Concatenation \\ method}} & \textbf{\makecell{Test \\ acc.}} & \textbf{\makecell{Test \\ F1}} \\
	\hline
	\textbf{\makecell{LMS \\ epochs: 100}} & *  & 55.20\% & 58.82\% \\
	\hline
	\textbf{\makecell{RMS \\ epochs: 100}} & *  & 56.50\% & 59.61\% \\
	\hline
	\textbf{\makecell{LMS + FFNN \\ epochs: 15}} & [$u,v,-u\oplus_Mv, d(u,v)$]   & 97.60\% & 97.58\% \\
	\hline
	\textbf{\makecell{RMS+FFNN \\  epochs: 15}} & [$u,v,-u\oplus_Mv, \cos(u,v), d(u,v)$]  & 97.50\% & 97.49\% \\
	\hline
	\textbf{\makecell{LMS + FFNN$_{c = 0.1}$ \\ epochs: 30}} & [$u,v,-u\oplus_Mv, d(u,v)$]  & 97.70\% & 97.70\% \\
	\hline
	\textbf{\makecell{RMS+FFNN$_{c = 0.1}$ \\  epochs: 30}} & [$u,v,-u\oplus_Mv, d(u,v)$]  & 97.80\% & 97.79\% \\
	\hline
	\textbf{\makecell{EA+FFNN \\ epochs: 15}} & [$u,v,|u-v|, u*v, \langle u , v \rangle, euclid \_ d(u,v)$]  & 66.70\% & 56.92\% \\
	\hline
	\textbf{\makecell{LSTM+FFNN \\ epochs: 100}} & [$u,v,|u-v|, u*v$]  & 96.90\% & 96.83\% \\
	\hline
	\textbf{\makecell{OE \\ epochs: 15}} &  *  & \textbf{99.20\%} & 99.40\% \\
	\hline
	\end{tabular}
	\captionof{table}{4-digit numbers dataset 2 class, FFNN: 256, embedding dimension: 50} 
	\label{table:numbers_results_embdim_50}
\end{center}

\begin{center} 
	\begin{tabular}{|c|c|c|c|c|c|}
	\hline
	\textbf{Model} & \textbf{\makecell{Concatenation \\ method}} & \textbf{\makecell{Test \\ acc.}} & \textbf{\makecell{Test \\ F1}} \\
	\hline
	\textbf{\makecell{LMS \\ epochs: 100}} & * & 53.30\% & 57.27\% \\
	\hline
	\textbf{\makecell{RMS \\ epochs: 100}} & * & 54.70\% & 61.77\% \\
	\hline
	\textbf{\makecell{LMS+FFNN \\ epochs: 15}} & [$u,v,-u\oplus_Mv, d(u,v)$]  & 97.80\% & 97.79\% \\
	\hline
	\textbf{\makecell{RMS+FFNN \\ epochs: 15}} & [$u,v,-u\oplus_Mv, \cos(u,v), d(u,v)$]  & 97.60\% & 97.60\% \\
	\hline
	\textbf{\makecell{LMS + FFNN$_{c = 0.1}$ \\ epochs: 30}} & [$u,v,-u\oplus_Mv, d(u,v)$]   & 96.10\% & 96.00\% \\
	\hline
	\textbf{\makecell{RMS+FFNN$_{c = 0.1}$ \\  epochs: 30}} & [$u,v,-u\oplus_Mv, \cos(u,v), d(u,v)$]  & 93.30\% & 92.85\% \\
	\hline
	\textbf{\makecell{EA+FFNN \\ epochs: 15}} & [$u,v,|u-v|, u*v$]  & 65.70\% & 53.33\% \\
	\hline
	\textbf{\makecell{LSTM+FFNN \\ epochs: 100}} & [$u,v,|u-v|, u*v$]  & 96.70\% & 96.62\% \\
	\hline
	\textbf{\makecell{OE \\ epochs: 15}} & *  & \textbf{99.10\%} & 98.42\% \\
	\hline
	
	\end{tabular}
	\captionof{table}{4-digit numbers 2 class, FFNN: 256, embedding dimension: 5} 
	\label{table:numbers_results_embdim_5}
\end{center}

We can also see the Euclidean Averaging model isn't able to effectively learn the "less than" operation for 4-digit numbers. This is expectable, since when we use euclidean averaging as a sentence representation method, we lose the word order. This doesn't occur with the \textbf{LMS+FFNN} and \textbf{RMS+FFNN} models, because of the non-commutativity and non-associativity of the mobius addition operation. However, the euclidean model is still better than random because, apart from having the FFNN, there are some instances where the model is able to make guesses better than random. 

For example if we have a sample such that the first number is constructed from the digits $\{1, 9, 9, 9\}$  and the digits $\{1, 1, 1, 2\}$ form the second number, although we couldn't claim that the first number is bigger than the second one, if we arrange the digits randomly there is a higher probability that the first number will end up being bigger.

Unlike the performance observed in Section \ref{section:experiments_adjnoun} here we can see that the models \textbf{LMS} and \textbf{RMS} aren't able to perform much better than random. This is due to the fact that there isn't an effective embeddings representation that can be learned using this simplistic approach. And unlike the \textbf{EA+FFNN} method, these methods do not have a FFNN on top to be able to perform at least visibly better than random.
	
For LSTMs, we ran 100 epochs for different LSTM cell sizes: 512,256,128 and report the best results.
For Order Embeddings we varied the GRU dimensions for 1024,512,256,128.

The superiority of the Order Embeddings model, is not a surprise given its complex architecture which uses recurrent networks (with a GRU unit).
It is interesting to see that the embedding dimension for the methods \textbf{LMS+FFNN} and \textbf{RMS+FFNN}, unlike in the experiments for SNLI and SICK, doesn't play a significant role. This is probably due to the fact that we operate with only 10 embeddings and even with a vector space of a small dimensionality of 5, there is enough space for positioning the embeddings such that a relatively good accuracy is achieved.

\begin{figure}
\begin{subfigure}{.45\textwidth}
  \centering
  \includegraphics[width=1.0\linewidth]{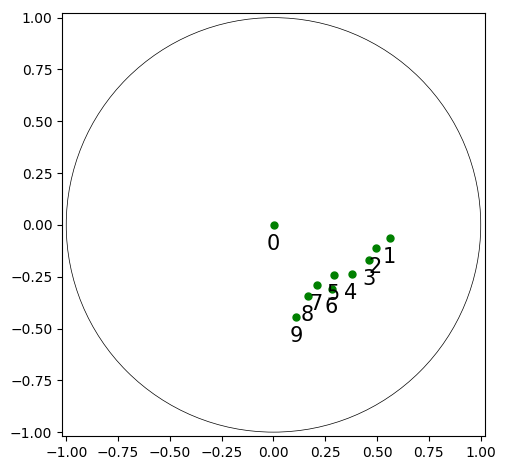}
  \caption{Epoch 1}
  \label{subfig:vis_numbers_epoch_1}
\end{subfigure}%
\hspace{0.1\textwidth}
\begin{subfigure}{.45\textwidth}
  \centering
  \includegraphics[width=1.0\linewidth]{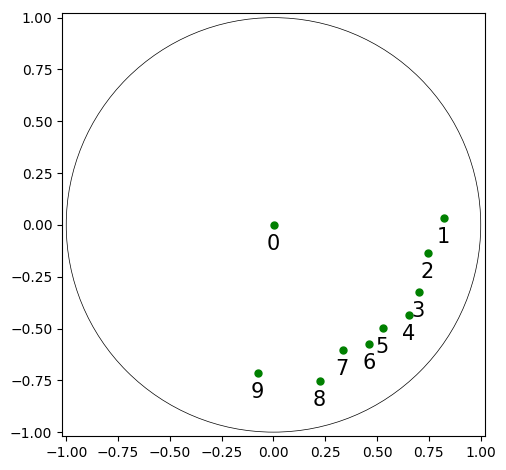}
  \caption{Epoch 5}
  \label{subfig:vis_numbers_epoch_5}
\end{subfigure}

\begin{subfigure}{.45\textwidth}
  \centering
  \includegraphics[width=1.0\linewidth]{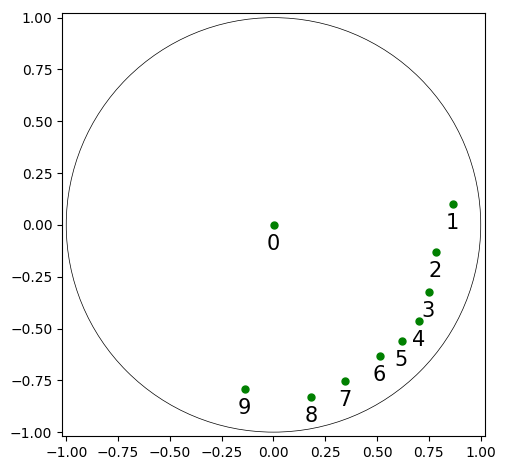}
  \caption{Epoch 10}
  \label{subfig:vis_numbers_epoch_10}
  
\end{subfigure}%
\hspace{0.1\textwidth}
\begin{subfigure}{.45\textwidth}
  \centering
  \includegraphics[width=1.0\linewidth]{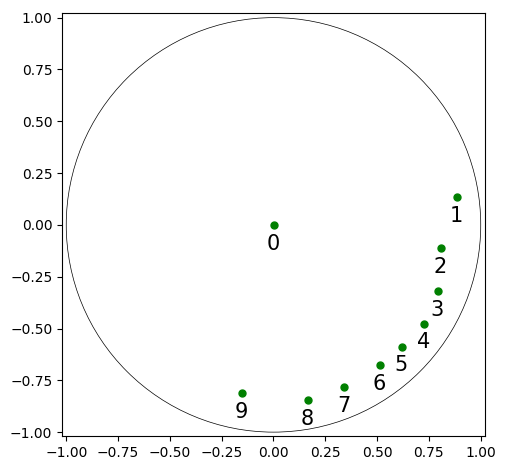}
  \caption{Epoch 15}
  \label{subfig:vis_numbers_epoch_15}
\end{subfigure}

\caption{Visualization for the training process of the \textbf{LMS+FFNN} method on the 4-digit numbers dataset. A test accuracy of 97.20\% was achieved even in this experiment with an embedding dimensionality 2.}
\label{fig:visualization_numbers_lms_ffnn}
\end{figure}

In Figure \ref{fig:visualization_numbers_lms_ffnn} we can see the training visualization for the \textbf{LMS+FFNN} method.

\section{Adjective-Noun}\label{section:experiments_adjnoun}

Here we present the results for the dataset introduced in Section \ref{section:datasets_adjnoun}. Both for \textbf{OE} and \textbf{LSTM+FFNN} we use 128 as a GRU cells size and LSTM cell size, respectively.
\begin{center} 
	\begin{tabular}{|c|c|c|c|c|c|}
	\hline
	\textbf{Model} & \textbf{\makecell{Concatenation \\ method}} & \textbf{\makecell{Test \\ acc.}} & \textbf{\makecell{Test \\ F1}} \\
	\hline
	\textbf{\makecell{LMS\\ epochs: 15}} & *  & \textbf{99.99\%} & 99.99\% \\
	\hline
	\textbf{\makecell{RMS \\ epochs: 15}} & * & 99.98\% & 99.97\% \\
	\hline
	\textbf{\makecell{LMS+FFNN \\ epochs: 70}} & [$u,v,-u\oplus_Mv, \cos(u,v), d(u,v)$] & 95.63\% & 95.64\% \\
	\hline
	\textbf{\makecell{RMS+FFNN \\ epochs: 70}} & [$u,v,-u\oplus_Mv, \cos(u,v), d(u,v)$] & 96.93\% & 96.92\% \\
	\hline
	\textbf{\makecell{LMS+FFNN$_{c = 0.01}$ \\ epochs: 70}} & [$u,v,-u\oplus_Mv, \cos(u,v), d(u,v)$] & 96.96\% & 96.92\% \\
	\hline
	\textbf{\makecell{RMS+FFNN$_{c = 0.03}$ \\ epochs: 70}} & [$u,v,-u\oplus_Mv, \cos(u,v), d(u,v)$] & 96.80\% & 96.81\% \\
	\hline
	\textbf{\makecell{EA+FFNN \\ epochs: 100}} & [$u,v,|u-v|, u*v, euclid \_ d(u,v)$] & 79.72\% & 82.50\% \\
	\hline
	\textbf{\makecell{LSTM+FFNN \\ epochs: 150}} & [$u,v,|u-v|, u*v$] & 87.12\% & 87.80\% \\
	\hline
	\textbf{\makecell{OE \\ epochs: 100}} & * & 99.95\% & 99.95\% \\
	\hline
	\end{tabular}
	\captionof{table}{Adjective-Noun dataset 2 class, FFNN: 256, embedding dimension: 50} 
	\label{table:adjnoun_results_report_embdim_50}
\end{center}

\begin{center} 
	\begin{tabular}{|c|c|c|c|c|c|c|}
	\hline
	\textbf{Model} & \textbf{\makecell{Concatenation \\ method}} & \textbf{\makecell{Test \\ acc.}} & \textbf{\makecell{Test \\ F1}} \\
	\hline
	\textbf{\makecell{LMS \\ epochs: 15}} & *  & 99.37\% & 99.37\% \\
	\hline
	\textbf{\makecell{RMS \\ epochs: 15}} & *  & \textbf{99.40\%} & 99.40\% \\
	\hline
	\textbf{\makecell{LMS+FFNN \\ epochs: 70}} & [$u,v,-u\oplus_Mv, d(u,v)$]  & 91.78\% & 91.95\% \\
	\hline
	\textbf{\makecell{RMS+FFNN \\  epochs: 70}} & [$u,v,-u\oplus_Mv, \cos(u,v), d(u,v)$]   & 80.64\% & 79.59\% \\
	\hline
	\textbf{\makecell{LMS+FFNN$_{c = 0.01}$ \\ epochs: 70}} & [$u,v,-u\oplus_Mv, \cos(u,v), d(u,v)$]  & 86.11\% & 86.66\% \\
	\hline
	\textbf{\makecell{RMS+FFNN$_{c = 0.03}$ \\ epochs: 70}} & [$u,v,-u\oplus_Mv, \cos(u,v), d(u,v)$]   & 86.76\% & 86.18\% \\
	\hline
	\textbf{\makecell{EA+FFNN \\ epochs: 100}} & [$u,v,|u-v|, u*v, euclid \_ d(u,v)$]  & 77.46\% & 80.71\% \\
	\hline
	\textbf{\makecell{LSTM+FFNN \\ epochs: 150}} & [$u,v,|u-v|, u*v$]  & 89.95\% & 90.83\% \\
	\hline
	\textbf{\makecell{OE \\ epochs: 100}} & * & 99.23\% & 99.09\% \\
	\hline
	\end{tabular}
	\captionof{table}{Adjective-Noun dataset 2 class, FFNN: 256, embedding dimension: 5} 
	\label{table:adjnoun_results_report_embdim_5}
\end{center}

\begin{figure}
\begin{subfigure}{.45\textwidth}
  \centering
  \includegraphics[width=1.0\linewidth]{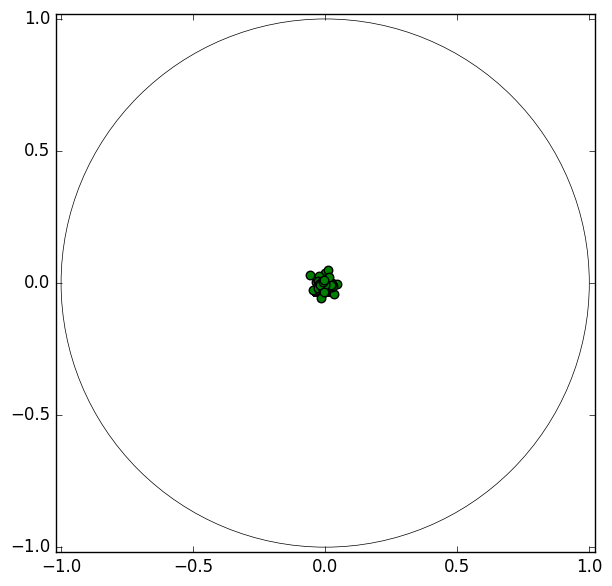}
  \caption{Epoch 1}
  \label{fig:vis_adj_noun_epoch_1}
\end{subfigure}%
\hspace{0.1\textwidth}
\begin{subfigure}{.45\textwidth}
  \centering
  \includegraphics[width=1.0\linewidth]{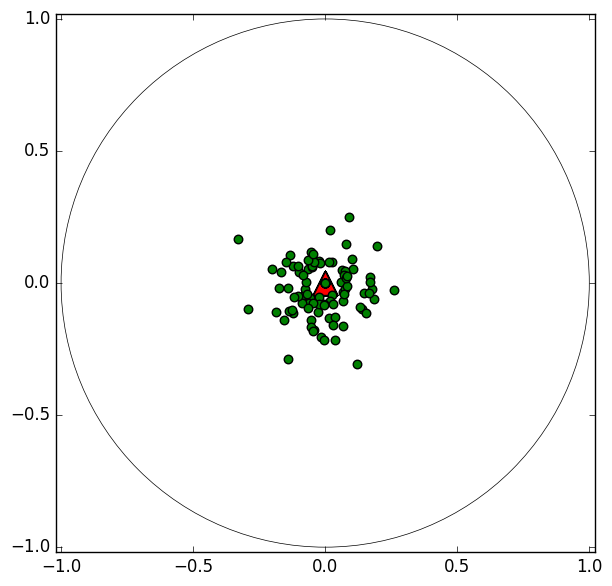}
  \caption{Epoch 5}
  \label{fig:vis_adj_noun_epoch_5}
\end{subfigure}

\begin{subfigure}{.45\textwidth}
  \centering
  \includegraphics[width=1.0\linewidth]{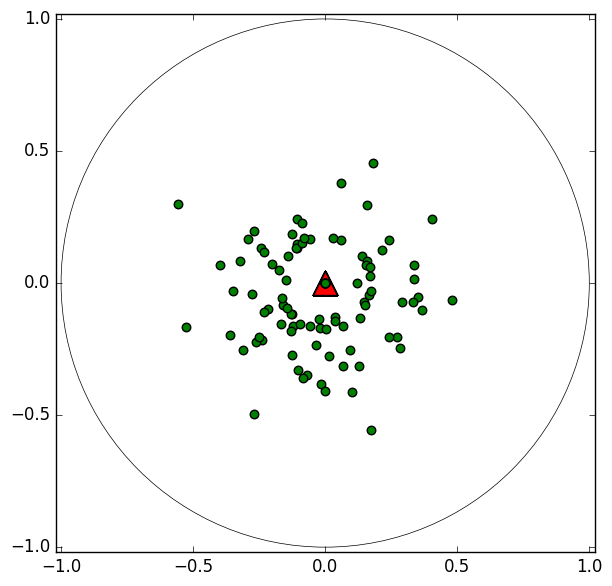}
  \caption{Epoch 10}
  \label{fig:vis_adj_noun_epoch_10}
  
\end{subfigure}%
\hspace{0.1\textwidth}
\begin{subfigure}{.45\textwidth}
  \centering
  \includegraphics[width=1.0\linewidth]{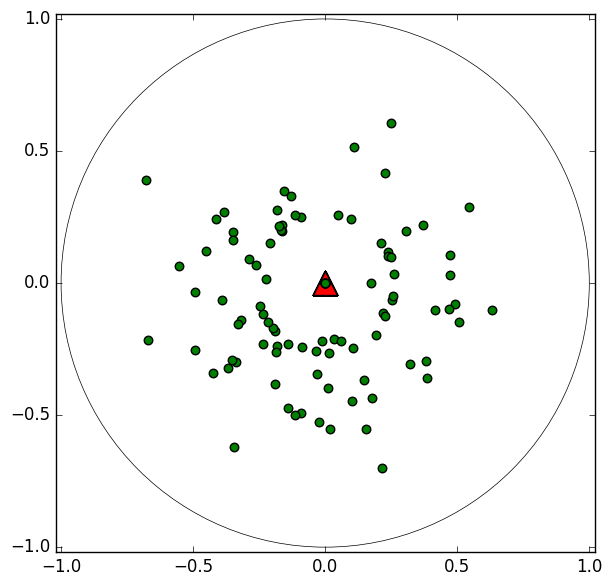}
  \caption{Epoch 15}
  \label{fig:vis_adj_noun_epoch_15}
\end{subfigure}

\begin{subfigure}{.45\textwidth}
  \centering
  \includegraphics[width=1.0\linewidth]{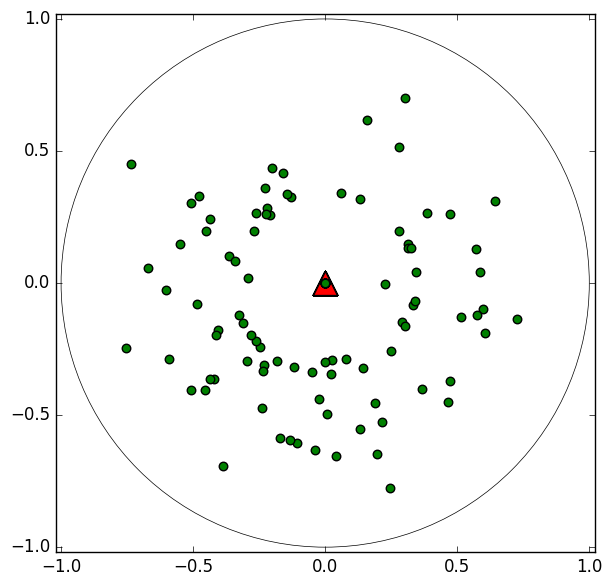}
  \caption{Epoch 20}
  \label{fig:vis_adj_noun_epoch_20}
  
\end{subfigure}%
\hspace{0.1\textwidth}
\begin{subfigure}{.45\textwidth}
  \centering
  \includegraphics[width=1.0\linewidth]{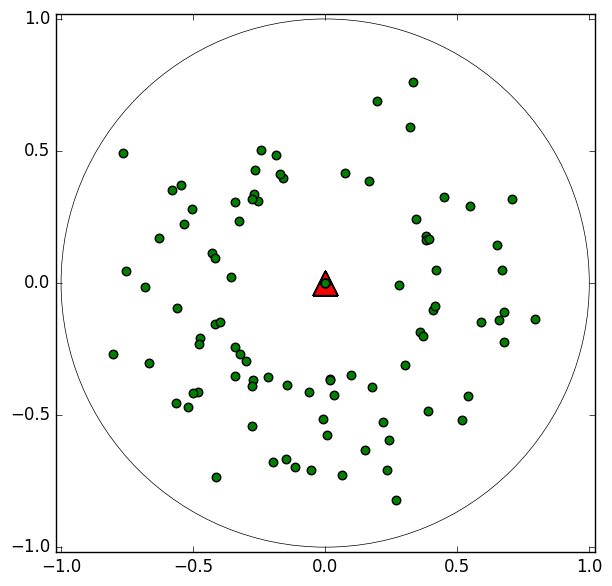}
  \caption{Epoch 25}
  \label{fig:vis_adj_noun_epoch_25}
\end{subfigure}

\caption{Visualization for the training process of the \textbf{LMS} method on the Adjective-Noun dataset. The green points represent nouns, whereas the red triangle represents the adjectevis which congregate around the origin. For visualization purposes we randomly picked a subset of adjectives and nouns and show them in this. An accuracy of 99.2\% was achieved even in this experiment with an embedding dimensionality 2}
\label{fig:visualization_adjnoun_lms}
\end{figure}

From Tables \ref{table:adjnoun_results_report_embdim_50} and \ref{table:adjnoun_results_report_embdim_5} we can see that the best results are achieved using the \textbf{LMS} and \textbf{RMS} methods. Although these methods are the least complex ones, they manage to learn very effective embedding representations, as it is shown in Figure \ref{fig:visualization_adjnoun_lms}. Namely, most of the adjective embeddings tend to congregate around the origin, whereas the noun embeddings tend to go further away from the origin, with each trained epoch. Recognizing an entailment pair of the form "$noun_i$ entails $adjective_j noun_i$" boils down to detecting whether the embeddings for the premise and hyptohesis are almost the same, since the adjectives tend be very close to the origin and do not influence the hypothesis representation using the mobius addition $adjective_j \oplus_M noun_i$.

\section{Parenthesization effect}\label{section:experiments_left_vs_right}
In this section we examine the importance of the sentence parse tree and the effect it has on the problem of Textual Entailment. We evaluate on the SICK and the SNLI dataset. 
\begin{center} 
	\begin{adjustwidth}{-0cm}{}
	\begin{tabular}{|c|c|c|c|c|}
	\hline
	\textbf{Model} & \textbf{\makecell{Concatenation \\ method}} & \textbf{\makecell{Test \\ acc.}} & \textbf{\makecell{Test \\ F1}} \\
	\hline
	\textbf{\makecell{LMS+FFNN}} & [$u,v,-u\oplus_Mv, \cos(u,v), d(u,v)$] & 85.98\% & 76.05\% \\
	\hline
	\textbf{\makecell{LMS+FFNN}} & [$u,v,-u\oplus_Mv, d(u,v)$] & 86.34\% & 75.32\% \\
	\hline
	\textbf{\makecell{RMS+FFNN}} & [$u,v,-u\oplus_Mv, \cos(u,v), d(u,v)$] & \textbf{86.50\%} & 76.54\% \\
	\hline
	\textbf{\makecell{RMS+FFNN}} & [$u,v,-u\oplus_Mv, d(u,v)$] & 85.65\%  & 74.87\% \\
	\hline
	\end{tabular}
	\end{adjustwidth}
	\captionof{table}{SICK dataset 2 class, embedding dim: 50, FFNN: 256, epochs 70} 
	\label{table:left_right_sick_50}
\end{center}

\begin{center} 
	\begin{adjustwidth}{-0cm}{}
	\begin{tabular}{|c|c|c|c|c|}
	\hline
	\textbf{Model} & \textbf{\makecell{Concatenation \\ method}} & \textbf{\makecell{Test \\ acc.}} & \textbf{\makecell{Test \\ F1}} \\
	\hline
	\textbf{\makecell{LMS+FFNN}} & [$u,v,-u\oplus_Mv, \cos(u,v), d(u,v)$] & 84.98\% & 74.66\% \\
	\hline
	\textbf{\makecell{LMS+FFNN}} & [$u,v,-u\oplus_Mv, d(u,v)$] & 84.72\%  & 72.73\% \\
	\hline
	\textbf{\makecell{RMS+FFNN}} & [$u,v,-u\oplus_Mv, \cos(u,v), d(u,v)$] & 85.29\% & 74.43\% \\
	\hline
	\textbf{\makecell{RMS+FFNN}} & [$u,v,-u\oplus_Mv, d(u,v)$] & \textbf{85.43\%} & 73.21\% \\
	\hline
	\end{tabular}
	\end{adjustwidth}
	\captionof{table}{SICK dataset 2 class, embedding dim: 5, FFNN: 256, epochs 70} 
	\label{table:left_right_sick_5}
\end{center}

\begin{center} 
	\begin{adjustwidth}{-0cm}{}
	\begin{tabular}{|c|c|c|c|c|c|}
	\hline
	\textbf{Model} & \textbf{\makecell{Concatenation \\ method}} & \textbf{\makecell{Val \\ acc.}} & \textbf{\makecell{Test \\ acc.}} & \textbf{\makecell{Test \\ F1}} \\
	\hline
	\textbf{\makecell{LMS+FFNN \\ epochs: 10}} & [$u,v,-u\oplus_Mv, \cos(u,v), d(u,v)$] & 83.67\% & \textbf{83.16\%} & 75.54\% \\
	\hline
	\textbf{\makecell{LMS+FFNN \\ epochs: 10}} & [$u,v,-u\oplus_Mv, d(u,v)$] & 83.41\% & 82.83\% & 74.73\% \\
	\hline
	\textbf{\makecell{RMS+FFNN\\ epochs: 3}} & [$u,v,-u\oplus_Mv, \cos(u,v), d(u,v)$] & 82.85\% & 82.91\% & 76.39\% \\
	\hline
	\textbf{\makecell{RMS+FFNN\\ epochs: 3}} & [$u,v,-u\oplus_Mv, d(u,v)$] & 81.97\% & 82.01\% & 73.89\% \\
	\hline
	\end{tabular}
	\end{adjustwidth}
	\captionof{table}{SNLI dataset 2 class, embedding dim: 50, FFNN: 256, epochs 10} 
	\label{table:left_right_snli_50}
\end{center}

\begin{center} 
	\begin{adjustwidth}{-0cm}{}
	\begin{tabular}{|c|c|c|c|c|c|}
	\hline
	\textbf{Model} & \textbf{\makecell{Concatenation \\ method}} & \textbf{\makecell{Val \\ acc.}} & \textbf{\makecell{Test \\ acc.}} & \textbf{\makecell{Test \\ F1}} \\
	\hline
	\textbf{\makecell{LMS+FFNN \\ epochs: 10}} & [$u,v,-u\oplus_Mv, \cos(u,v), d(u,v)$] & 81.20\% & 80.80\% & 73.07\% \\
	\hline
	\textbf{\makecell{LMS+FFNN \\ epochs: 10}} & [$u,v,-u\oplus_Mv, d(u,v)$] & 81.41\% & \textbf{81.46\%} & 72.91\% \\
	\hline
	\textbf{\makecell{RMS+FFNN\\ epochs: 5}} & [$u,v,-u\oplus_Mv, \cos(u,v), d(u,v)$] & 81.39\% & 80.71\% & 72.23\% \\
	\hline
	\textbf{\makecell{RMS+FFNN\\ epochs: 5}} & [$u,v,-u\oplus_Mv, d(u,v)$] & 81.40\% & 80.88\% & 71.29\% \\
	\hline
	\end{tabular}
	\end{adjustwidth}
	\captionof{table}{SNLI dataset 2 class, embedding dim: 5, FFNN: 256, epochs 10} 
	\label{table:left_right_snli_5}
\end{center}

If we compare the results from Tables \ref{table:left_right_sick_50} and \ref{table:left_right_sick_5} to the results in Tables \ref{table:sick_results_report_embdim_50_class_2} and \ref{table:sick_results_report_embdim_5_class_2}, respectively, we can see that the parse tree in the \textbf{MS+FFNN} method doesn't add much value in comparison with left or right parenthesization. We arrive at the same conclusion if we compare the results obtained on the SNLI dataset in Tables \ref{table:left_right_snli_50} and \ref{table:left_right_snli_5} in comparison to the results in Tables \ref{table:snli_results_report_embdim_50_class_2} and \ref{table:snli_results_report_embdim_5_class_2}. An intuitive explanation could be the fact that when using the approach described in Section \ref{section:sentence_representation}, although mobius addition is non-commutative and non-associative, a much bigger factor is the positioning of the word embeddings in the Poincare ball than their respective order that is used to sum them up.

\section{Disentanglement}\label{section:experiments_disentanglement}
Similarly to what was done in \cite{poincare} in Section 4.1, we use the loss function defined as: 
\begin{equation} \label{eq:loss_disentanglement}
\Loss= -\sum_{(u,v) \in D}\log\frac{\exp(-d(u,v))}{\sum_{(u',v') \in S}\exp(-d(u',v')) + \exp(-d(u,v))}
\end{equation}

where $S$ represents a random set of negative samples. $u,v$ represent the respective sentence embeddings and $d(u,v)$ represents the Poincare distance.

We use this loss function for disentangling the word embeddings in the first 3 epochs of the training phase, and then switch to the standard loss function with FFNN as defined in Section \ref{section:mobius_summation_ffnn}. The expectations were that starting with one loss function would help with the initial positioning of the embeddings. Although suitable for Link prediction as reported in \cite{poincare}, this doesn't generalize well to sentence embeddings since the word embeddings tend to shrink near the origin. Consecutively, the loss function, defined in Equation \ref{eq:loss_disentanglement}, converges to zero without achieving the wanted effect of embeddings disentanglement.

\section{Curvatures experiments} \label{section:experiments_curvature}
In this section we focus on showing the relation between the Euclidean space and the Poincare hyperbolic space as we change the $c$ curvature factor that was introduced in Section \ref{section:curvatures}. As we can see from the Tables \ref{table:curvature_sick_class_2_emb_50}, \ref{table:curvature_sick_class_2_emb_5}, \ref{table:curvature_sick_class_3_emb_50}, \ref{table:curvature_sick_class_3_emb_5} when $c$ becomes very small we tend to get results very similar results to those when using Euclidean Summation, which we define as \textbf{ES+FFNN}. 

\begin{center} 
	\begin{adjustwidth}{-0cm}{}
	\begin{tabular}{|c|c|c|c|c|c|}
	\hline
	\textbf{Model} & \textbf{\makecell{Concatenation \\ method}} & \textbf{\makecell{Test \\ acc.}} & \textbf{\makecell{Test \\ F1}} \\
	\hline
	\textbf{\makecell{MS+FFNN$_{c=1.0}$}} & [$u,v,|-u\oplus_Mv|$] & 84.80\% & 72.95\% \\
	\hline
	\textbf{\makecell{MS+FFNN$_{c=0.1}$}} & [$u,v,|-u\oplus_Mv|$] & 85.08\% & 75.70\% \\
	\hline
	\textbf{\makecell{MS+FFNN$_{c=0.03}$}} & [$u,v,|-u\oplus_Mv|$] & 84.72\% & 74.08\% \\
	\hline
	\textbf{\makecell{MS+FFNN$_{c=0.01}$}} & [$u,v,|-u\oplus_Mv|$] & 84.98\% & 74.92\% \\
	\hline
	\textbf{\makecell{MS+FFNN$_{c=0.001}$}} & [$u,v,|-u\oplus_Mv|$] & 85.71\% & 74.95\% \\
	\hline
	\textbf{\makecell{MS+FFNN$_{c=0.0001}$}} & [$u,v,|-u\oplus_Mv|$] & 84.80\% & 72.75\% \\
	\hline
	\textbf{\makecell{MS+FFNN$_{c=0.000001}$}} & [$u,v,|-u\oplus_Mv|$] & 85.26\% & 73.45\% \\
	\hline
	\textbf{\makecell{MS+FFNN$_{c=0.00000001}$}} & [$u,v,|-u\oplus_Mv|$] & 85.33\% & 75.13\% \\
	\hline
	\textbf{\makecell{ES+FFNN}} & [$u,v,|u-v|$] & 85.69\% & 74.90\% \\
	\hline
	\end{tabular}
	\end{adjustwidth}
	\captionof{table}{SICK dataset 2 class, embedding dim: 50, FFNN: 256, epochs 100} 
	\label{table:curvature_sick_class_2_emb_50}
\end{center}

\begin{center} 
	\begin{adjustwidth}{-0cm}{}
	\begin{tabular}{|c|c|c|c|c|c|}
	\hline
	\textbf{Model} & \textbf{\makecell{Concatenation \\ method}} & \textbf{\makecell{Test \\ acc.}} & \textbf{\makecell{Test \\ F1}} \\
	\hline
	\textbf{\makecell{MS+FFNN$_{c=1.0}$}} & [$u,v,|-u\oplus_Mv|$] & 84.84\% & 73.78\% \\
	\hline
	\textbf{\makecell{MS+FFNN$_{c=0.1}$}} & [$u,v,|-u\oplus_Mv|$] & 82.85\% & 71.07\% \\
	\hline
	\textbf{\makecell{MS+FFNN$_{c=0.03}$}} & [$u,v,|-u\oplus_Mv|$]  & 81.53\% & 61.41\% \\
	\hline
	\textbf{\makecell{MS+FFNN$_{c=0.01}$}} & [$u,v,|-u\oplus_Mv|$]  & 83.60\% & 70.62\% \\
	\hline
	\textbf{\makecell{MS+FFNN$_{c=0.001}$}} & [$u,v,|-u\oplus_Mv|$]  & 82.95\% & 70.19\% \\
	\hline
	\textbf{\makecell{MS+FFNN$_{c=0.0001}$}} & [$u,v,|-u\oplus_Mv|$]  & 81.71\% & 68.17\% \\
	\hline
	\textbf{\makecell{MS+FFNN$_{c=0.000001}$}} & [$u,v,|-u\oplus_Mv|$] & 83.93\% & 70.20\% \\
	\hline
	\textbf{\makecell{MS+FFNN$_{c=0.00000001}$}} & [$u,v,|-u\oplus_Mv|$] & 83.74\% & 71.69\% \\
	\hline
	\textbf{\makecell{ES+FFNN}} & [$u,v,|u-v|$] & 83.66\% & 71.38\% \\
	\hline
	\end{tabular}
	\end{adjustwidth}
	\captionof{table}{SICK dataset 2 class, embedding dim: 5, FFNN: 256, epochs 100} 
	\label{table:curvature_sick_class_2_emb_5}
\end{center}

\begin{center} 
	\begin{adjustwidth}{-0cm}{}
	\begin{tabular}{|c|c|c|c|c|c|}
	\hline
	\textbf{Model} & \textbf{\makecell{Concatenation \\ method}} & \textbf{\makecell{Test \\ acc.}} \\
	\hline
	\textbf{\makecell{MS+FFNN$_{c=1.0}$}} & [$u,v,|-u\oplus_Mv|$]  & 	80.47\%	\\
	\hline
	\textbf{\makecell{MS+FFNN$_{c=0.1}$}} & [$u,v,|-u\oplus_Mv|$]  & 	81.14\%	\\
	\hline
	\textbf{\makecell{MS+FFNN$_{c=0.03}$}} & [$u,v,|-u\oplus_Mv|$]  & 	79.93\%	\\
	\hline
	\textbf{\makecell{MS+FFNN$_{c=0.01}$}} & [$u,v,|-u\oplus_Mv|$]  & 	81.02\%	\\
	\hline
	\textbf{\makecell{MS+FFNN$_{c=0.001}$}} & [$u,v,|-u\oplus_Mv|$]  &  81.17\%	\\
	\hline
	\textbf{\makecell{MS+FFNN$_{c=0.0001}$}} & [$u,v,|-u\oplus_Mv|$]  &  79.40\%	\\
	\hline
	\textbf{\makecell{MS+FFNN$_{c=0.000001}$}} & [$u,v,|-u\oplus_Mv|$] &  79.72\%	\\
	\hline
	\textbf{\makecell{MS+FFNN$_{c=0.00000001}$}} & [$u,v,|-u\oplus_Mv|$] &  81.00\% \\
	\hline
	\textbf{\makecell{ES+FFNN}} & [$u,v,|u-v|$] & 80.74\% \\
	\hline
	\end{tabular}
	\end{adjustwidth}
	\captionof{table}{SICK dataset 3 class, embedding dim: 50, FFNN: 256, epochs 100} 
	\label{table:curvature_sick_class_3_emb_50}
\end{center}

\begin{center} 
	\begin{adjustwidth}{-0cm}{}
	\begin{tabular}{|c|c|c|c|c|c|}
	\hline
	\textbf{Model} & \textbf{\makecell{Concatenation \\ method}} & \textbf{\makecell{Test \\ acc.}} \\
	\hline
	\textbf{\makecell{MS+FFNN$_{c=1.0}$}} & [$u,v,|-u\oplus_Mv|$]  & 	76.01\%	\\
	\hline
	\textbf{\makecell{MS+FFNN$_{c=0.1}$}} & [$u,v,|-u\oplus_Mv|$]  & 	80.33\%	\\
	\hline
	\textbf{\makecell{MS+FFNN$_{c=0.03}$}} & [$u,v,|-u\oplus_Mv|$]  & 	78.02\%	\\
	\hline
	\textbf{\makecell{MS+FFNN$_{c=0.01}$}} & [$u,v,|-u\oplus_Mv|$]  & 	77.69\%	\\
	\hline
	\textbf{\makecell{MS+FFNN$_{c=0.001}$}} & [$u,v,|-u\oplus_Mv|$]  &  78.16\%	\\
	\hline
	\textbf{\makecell{MS+FFNN$_{c=0.0001}$}} & [$u,v,|-u\oplus_Mv|$]  &  79.89\%	\\
	\hline
	\textbf{\makecell{MS+FFNN$_{c=0.000001}$}} & [$u,v,|-u\oplus_Mv|$] &  79.50\%\\
	\hline
	\textbf{\makecell{MS+FFNN$_{c=0.00000001}$}} & [$u,v,|-u\oplus_Mv|$] & 78.97\% \\
	\hline
	\textbf{\makecell{ES+FFNN}} & [$u,v,|u-v|$] & 78.83\% \\
	\hline
	\end{tabular}
	\end{adjustwidth}
	\captionof{table}{SICK dataset 3 class, embedding dim: 5, FFNN: 256, epochs 100} 
	\label{table:curvature_sick_class_3_emb_5}
\end{center}

\begin{center} 
	\begin{adjustwidth}{-0cm}{}
	\begin{tabular}{|c|c|c|c|c|c|}
    	\hline
    	\textbf{Model} & \textbf{\makecell{Concatenation \\ method}} & \textbf{\makecell{Test \\ acc.}} \\
    	\hline
    	\textbf{\makecell{MS+FFNN$_{c=0.1}$}} & [$u,v,-u\oplus_Mv, \cos(u,v), d(u,v)$] & $85.0800  \pm 0.1327 \%$	\\
    	\hline
    	\textbf{\makecell{MS+FFNN$_{c=0.01}$}} & [$u,v,-u\oplus_Mv, \cos(u,v), d(u,v)$]  & $85.4300  \pm 0.1316 \%$	\\
    	\hline
    	\textbf{\makecell{MS+FFNN$_{c=0.03}$}} & [$u,v,-u\oplus_Mv, \cos(u,v), d(u,v)$]  & $85.0920  \pm 0.1320 \%$	\\
    	\hline
    	\textbf{\makecell{MS+FFNN$_{c=0.1}$}} & [$u,v,-u\oplus_Mv, d(u,v)$]  & $84.4580  \pm  0.2660 \%$	\\
    	\hline
    	\textbf{\makecell{MS+FFNN$_{c=0.01}$}} & [$u,v,-u\oplus_Mv, d(u,v)$]  & $85.1240  \pm  0.1570 \%$	\\
    	\hline
    	\textbf{\makecell{MS+FFNN$_{c=0.03}$}} & [$u,v,-u\oplus_Mv, d(u,v)$]  & $84.6800  \pm 0.2826 \%$	\\
    	\hline
	\end{tabular}
	\end{adjustwidth}
	\captionof{table}{SICK dataset 3 class, embedding dim: 5, FFNN: 256, epochs 100} 
	\label{table:curvature_sick_class_3_emb_5}
\end{center}

Finally, what we can conclude is that there is no clear pattern to which value of $c$ performs the best, but it is definitely indicative that the optimal space for the embeddings is somewhere between a Poincare unit ball (when $c=1$) and a Euclidean vector space ($c=0$).


%% file: conclusion.tex

\chapter{Conclusion}
\label{ch:conclusion}

In this thesis we have presented a completely novel way of embedding sentences with the aim of solving the problem of Textual entailment. With combining mobius addition in a hyperbolic space and exploiting the structure of the sentence we have shown the advantages and the limitations that our parameterless sentence composition method has over more traditional methods like LSTMs and Euclidean Averaging. We have also measured and analyzed the performances our methods have across different types of datasets. Although not beating the state of the art in this field on the SNLI dataset, our contribution consists of introducing a model of a completely different nature that requires very few parameters and convergence time. Outperforming all the other baselines on the SICK dataset shows the adaptability of the Mobius Summation model and how it is able to learn quickly with little data.

\section{Future work}

Although the premise of this work was that hyperbolic spaces are suitable for modeling data of hierarchical nature and that entailment can be defined as a hierarchy relationships between objects, we believe that the hyperbolic sentence representation introduced in this thesis can be applied to solving a variety of problems in the field of Natural Language Understanding.


%% file: thesis.bbl
\begin{thebibliography}{10}

\bibitem{snli_corpus}
S.~R. Bowman, G.~Angeli, C.~Potts, and C.~D. Manning, ``A large annotated corpus for learning natural language inference,'' {\em arXiv preprint arXiv:1508.05326}, 2015.

\bibitem{skipgram}
T.~Mikolov, I.~Sutskever, K.~Chen, G.~S. Corrado, and J.~Dean, ``Distributed representations of words and phrases and their compositionality,'' in {\em Advances in neural information processing systems}, pp.~3111--3119, 2013.

\bibitem{skipthoughts}
R.~Kiros, Y.~Zhu, R.~R. Salakhutdinov, R.~Zemel, R.~Urtasun, A.~Torralba, and S.~Fidler, ``Skip-thought vectors,'' in {\em Advances in neural information processing systems}, pp.~3294--3302, 2015.

\bibitem{a_simple_but_tough}
S.~Arora, Y.~Liang, and T.~Ma, ``A simple but tough-to-beat baseline for sentence embeddings,'' 2016.

\bibitem{fast_sent_jaggi}
M.~Pagliardini, P.~Gupta, and M.~Jaggi, ``Unsupervised learning of sentence embeddings using compositional n-gram features,'' {\em arXiv preprint arXiv:1703.02507}, 2017.

\bibitem{tensorflow}
M.~Abadi, P.~Barham, J.~Chen, Z.~Chen, A.~Davis, J.~Dean, M.~Devin, S.~Ghemawat, G.~Irving, M.~Isard, {\em et~al.}, ``Tensorflow: A system for large-scale machine learning.,'' in {\em OSDI}, vol.~16, pp.~265--283, 2016.

\bibitem{siamese}
T.~Kenter, A.~Borisov, and M.~de~Rijke, ``Siamese cbow: Optimizing word embeddings for sentence representations,'' {\em arXiv preprint arXiv:1606.04640}, 2016.

\bibitem{rnn_language_models}
T.~Mikolov, M.~Karafi{\'a}t, L.~Burget, J.~{\v{C}}ernock{\`y}, and S.~Khudanpur, ``Recurrent neural network based language model,'' in {\em Eleventh Annual Conference of the International Speech Communication Association}, 2010.

\bibitem{rnn_blog}
J.~Ma, ``All of recurrent neural networks.'' \url{https://medium.com/@jianqiangma/all-about-recurrent-neural-networks-9e5ae2936f6e}.
\newblock Accessed: 2018-01-04.

\bibitem{lstms}
S.~Hochreiter and J.~Schmidhuber, ``Long short-term memory,'' {\em Neural computation}, vol.~9, no.~8, pp.~1735--1780, 1997.

\bibitem{rocktaschel2015reasoning}
T.~Rockt{\"a}schel, E.~Grefenstette, K.~M. Hermann, T.~Ko{\v{c}}isk{\`y}, and P.~Blunsom, ``Reasoning about entailment with neural attention,'' {\em arXiv preprint arXiv:1509.06664}, 2015.

\bibitem{wang2015learning}
S.~Wang and J.~Jiang, ``Learning natural language inference with lstm,'' {\em arXiv preprint arXiv:1512.08849}, 2015.

\bibitem{liu2016learning}
Y.~Liu, C.~Sun, L.~Lin, and X.~Wang, ``Learning natural language inference using bidirectional lstm model and inner-attention,'' {\em arXiv preprint arXiv:1605.09090}, 2016.

\bibitem{nli_learning_scheme}
A.~Conneau, D.~Kiela, H.~Schwenk, L.~Barrault, and A.~Bordes, ``Supervised learning of universal sentence representations from natural language inference data,'' {\em arXiv preprint arXiv:1705.02364}, 2017.

\bibitem{order_embeddings}
I.~Vendrov, R.~Kiros, S.~Fidler, and R.~Urtasun, ``Order-embeddings of images and language,'' {\em arXiv preprint arXiv:1511.06361}, 2015.

\bibitem{poincare}
M.~Nickel and D.~Kiela, ``Poincar{\'e} embeddings for learning hierarchical representations,'' in {\em Advances in Neural Information Processing Systems}, pp.~6341--6350, 2017.

\bibitem{image_tangent_space}
Alexright, ``Tangent space image.'' \url{https://commons.wikimedia.org/wiki/File:Image_Tangent-plane.svg}.
\newblock Accessed: 2018-05-04.

\bibitem{kleinberg2007geographic}
R.~Kleinberg, ``Geographic routing using hyperbolic space,'' in {\em INFOCOM 2007. 26th IEEE International Conference on Computer Communications. IEEE}, pp.~1902--1909, IEEE, 2007.

\bibitem{krioukov2010hyperbolic}
D.~Krioukov, F.~Papadopoulos, M.~Kitsak, A.~Vahdat, and M.~Bogun{\'a}, ``Hyperbolic geometry of complex networks,'' {\em Physical Review E}, vol.~82, no.~3, p.~036106, 2010.

\bibitem{boguna2010sustaining}
M.~Bogun{\'a}, F.~Papadopoulos, and D.~Krioukov, ``Sustaining the internet with hyperbolic mapping,'' {\em Nature communications}, vol.~1, p.~62, 2010.

\bibitem{img_parallel_postulate}
D.~Didur, ``Non-euclidean geometry.'' \url{https://commons.wikimedia.org/wiki/File:Image_Tangent-plane.svg}.
\newblock Accessed: 2018-21-04.

\bibitem{poincare_dist_images}
J.~Jain, ``Implementing poincaré embeddings.'' \url{https://rare-technologies.com/implementing-poincare-embeddings/#}.
\newblock Accessed: 2018-01-04.

\bibitem{mobius}
G.~S. Birman and A.~A. Ungar, ``The hyperbolic derivative in the poincar{\'e} ball model of hyperbolic geometry,'' {\em Journal of mathematical analysis and applications}, vol.~254, no.~1, pp.~321--333, 2001.

\bibitem{bebis1994feed}
G.~Bebis and M.~Georgiopoulos, ``Feed-forward neural networks,'' {\em IEEE Potentials}, vol.~13, no.~4, pp.~27--31, 1994.

\bibitem{bottou2010large}
L.~Bottou, ``Large-scale machine learning with stochastic gradient descent,'' in {\em Proceedings of COMPSTAT'2010}, pp.~177--186, Springer, 2010.

\bibitem{riemannian_optimization}
S.~Bonnabel, ``Stochastic gradient descent on riemannian manifolds,'' {\em IEEE Transactions on Automatic Control}, vol.~58, no.~9, pp.~2217--2229, 2013.

\bibitem{flickr30}
P.~Young, A.~Lai, M.~Hodosh, and J.~Hockenmaier, ``From image descriptions to visual denotations: New similarity metrics for semantic inference over event descriptions,'' {\em Transactions of the Association for Computational Linguistics}, vol.~2, pp.~67--78, 2014.

\bibitem{sick_corpus}
M.~Marelli, L.~Bentivogli, M.~Baroni, R.~Bernardi, S.~Menini, and R.~Zamparelli, ``Semeval-2014 task 1: Evaluation of compositional distributional semantic models on full sentences through semantic relatedness and textual entailment,'' in {\em Proceedings of the 8th international workshop on semantic evaluation (SemEval 2014)}, pp.~1--8, 2014.

\bibitem{stanford_parser}
M.-C. De~Marneffe, B.~MacCartney, C.~D. Manning, {\em et~al.}, ``Generating typed dependency parses from phrase structure parses,'' in {\em Proceedings of LREC}, vol.~6, pp.~449--454, Genoa Italy, 2006.

\end{thebibliography}
